\documentclass[11pt,letterpaper]{article}

\usepackage{times}
\usepackage{fullpage}
\usepackage{microtype}
\usepackage{graphicx}
\usepackage{booktabs}
\usepackage{caption}
\usepackage{amsmath}
\usepackage{amssymb}
\usepackage{mathtools}
\usepackage{amsthm}
\usepackage{bm}

\usepackage{natbib}
\usepackage{xurl}
\usepackage{xcolor}
\usepackage[colorlinks,
            linkcolor=red,
            anchorcolor=blue,
            citecolor=blue]{hyperref}
\usepackage[capitalize,noabbrev,nameinlink]{cleveref}

\usepackage{titlesec}
\titleformat{\section}
  {\normalfont\large\bfseries}{\thesection}{1em}{}
\titleformat{\subsection}
  {\normalfont\normalsize\bfseries}{\thesubsection}{1em}{}
\titleformat{\subsubsection}
  {\normalfont\normalsize\bfseries}{\thesubsubsection}{1em}{}
\titlespacing*{\section}{0pt}{5pt plus 1pt minus 1pt}{2pt plus 1pt minus 1pt}
\titlespacing*{\subsection}{0pt}{4pt plus 1pt minus 1pt}{2pt plus 1pt minus 1pt}
\titlespacing*{\subsubsection}{0pt}{4pt plus 1pt minus 1pt}{2pt plus 1pt minus 1pt}
\titlespacing*{\paragraph}{0pt}{2pt plus 1pt minus 1pt}{4pt plus 1pt minus 1pt}

\renewenvironment{quote}
  {\list{}{\leftmargin=1em \rightmargin=1em}\item\relax}
  {\endlist}

\theoremstyle{plain}
\newtheorem{theorem}{Theorem}[section]
\newtheorem{proposition}[theorem]{Proposition}
\newtheorem{lemma}[theorem]{Lemma}

\newtheorem{conjecture}[theorem]{Conjecture}
\theoremstyle{definition}
\newtheorem{definition}[theorem]{Definition}

\usepackage{amsmath,amsfonts,bm}

\def\1{\bm{1}}

\def\valpha{{\bm{\alpha}}}

\def\vd{{\bm{d}}}

\def\vf{{\bm{f}}}

\def\vl{{\bm{l}}}

\def\vx{{\bm{x}}}
\def\vy{{\bm{y}}}

\def\evf{{f}}

\def\evl{{l}}

\def\evx{{x}}

\DeclareMathAlphabet{\mathsfit}{\encodingdefault}{\sfdefault}{m}{sl}
\SetMathAlphabet{\mathsfit}{bold}{\encodingdefault}{\sfdefault}{bx}{n}

\def\gA{{\mathcal{A}}}

\def\gC{{\mathcal{C}}}

\def\gG{{\mathcal{G}}}

\def\gL{{\mathcal{L}}}

\def\gP{{\mathcal{P}}}

\def\gT{{\mathcal{T}}}

\def\gV{{\mathcal{V}}}

\def\gY{{\mathcal{Y}}}

\newcommand{\R}{\mathbb{R}}

\newcommand{\conv}{\operatorname{conv}}

\title{\textbf{\Large{Goemans's Unsplittable-Flow Conjecture: The Anatomy of AI-Assisted Counterexample Discovery}}}
\author{
\normalsize
Jingwei Ge$^{1}$ \qquad
Na Xie$^{2}$ \qquad
Li Li$^{3}$ \qquad
Jiayang Li$^{1}$\thanks{Corresponding author. Email: \texttt{jiayangl@hku.hk}}
\\[5pt]
\small
$^{1}$The University of Hong Kong, Hong Kong SAR, China
\\
\small
$^{2}$Central University of Finance and Economics, Beijing, China
\\
\small
$^{3}$Tsinghua University, Beijing, China
}
\date{}

\begin{document}
\maketitle

\begin{abstract}
Goemans conjectured that any fractional single-source flow could be converted into an unsplittable flow without increasing total cost and while increasing any arc load by at most the largest demand. Recently, Dmitry Rybin announced a seven-vertex counterexample found through an interaction with ChatGPT, which disproved this long-standing conjecture. However, the interaction did not reveal why this particular topology and its demand, fractional-flow, and cost choices work together, whether the example belongs to a broader family, or which structural features determine where counterexamples can occur. It also left unclear why the AI-assisted search succeeded where earlier human efforts had not. In this paper, we develop a forward counterexample construction pipeline. It starts from an arborescence on which failure is impossible and adds terminal-entering shortcut arcs. The resulting load constraints define a finite set of admissible route choices; their convex hull, the conflict polytope, yields exact conditions for demands, fractional flows, and costs. Starting from a simple arborescence, the pipeline selects Rybin's topology and characterizes all counterexamples it supports. Experiments further demonstrate the pipeline's transferability: when instructed to follow it while varying the initial arborescence and shortcut placement, ChatGPT produced counterexamples on topologies genuinely different from Rybin's and found valid constructions starting from 93 other rooted arborescence shapes. The theory also explains why the original example was difficult to discover through blind enumeration and suggests that Rybin's interaction lies between blind enumeration and a complete theory-guided construction. 

\vspace{2em}
\vspace{1em}
\end{abstract}

\vspace{1em}

\section{Introduction}

Many freight distribution systems route shipments from a single source to multiple terminals. At the planning stage, allowing each demand to be divided among several paths yields a tractable fractional-flow relaxation. Operational implementation, however, may require an individual shipment to remain intact and follow a single path. This indivisibility requirement gives rise to the single-source unsplittable flow problem, a classic model in network optimization for studying whether a divisible flow plan can be converted into implementable single-path assignments \citep{Kleinberg1996,DinitzGargGoemans1999,KolliopoulosStein2002,Skutella2002}.

A fundamental theorem of \citet*{DinitzGargGoemans1999} shows that every fractional flow can be converted into an unsplittable flow while increasing the load on each arc by no more than the largest individual demand. This guarantee limits the additional arc load caused by enforcing indivisibility. A cost-preserving strengthening was subsequently conjectured by Goemans: the conversion should also be achievable without increasing total routing cost\footnote{\citet{SalazarSkutella2009} attribute this conjecture to a January 2000 personal communication from Goemans.}.  If true, the conjecture would imply that indivisible implementation requires only a sharply controlled additive load increase, with no deterioration in the objective value of the fractional plan. Over time, this conjecture has been proved in several restricted
settings \citep[e.g.,][]{Skutella2002,MorellSkutella2022,AlmoghrabiSkutellaWarode2026,TraubVargasKochZenklusen2026}.

\begin{figure}[ht]
\centering
\includegraphics[width=0.4\textwidth]{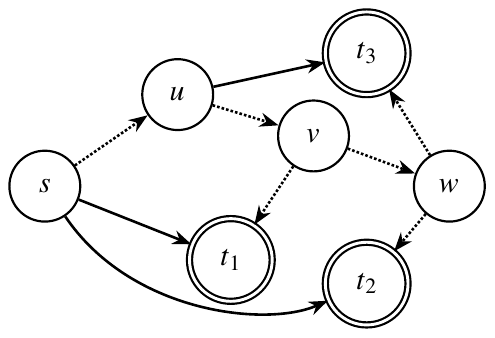}
\caption{Topology of Rybin's seven-vertex counterexample. The source is $s$; $u$, $v$, and $w$ are intermediate vertices; and $t_1$, $t_2$, and $t_3$ are terminals. Dotted arcs have zero unit cost, whereas solid arcs have positive unit cost.}
\label{fig:rybin-certificate}
\end{figure}

Recently, the long-standing conjecture abruptly fell. In July 2026, Dmitry Rybin publicly announced a counterexample produced through an interaction with ChatGPT \citep{Rybin2026}. The four prompts that led to the announced certificate are reproduced verbatim below:
\begin{quote}
Prompt 1: ``\textit{Construct a counterexample to general (non-planar) case of Dinitz Garg Goemans conjecture. You should do a breakthrough and find astructured counterexample.}''

\noindent
Prompt 2: ``\textit{Please continue research and find a complete unconditional counterexample.}''

\noindent
Prompt 3: ``\textit{Continue the search. Have a clear strategy obtained from deeper understanding of the problem structure.}''

\noindent
Prompt 4: ``\textit{It's enough of partial results. let's finish with a complete unconditional counterexample.}''
\end{quote}
\smallskip
The returned counterexample is almost disarmingly small. As shown in Figure~\ref{fig:rybin-certificate}, it has seven vertices --- one source $s$, three intermediate vertices $u$, $v$, and $w$, and three terminals $t_1$, $t_2$, and $t_3$ --- connected by nine arcs. Each terminal has exactly two source--terminal paths, so the complete unsplittable problem contains only $2^3=8$ route selections. The demands and positive unit arc costs are small integers, and the strict cost gap follows from three direct load comparisons. The cost structure is equally simple. Only the three solid arcs have positive unit costs. Once the topology and numerical data are given, the entire counterexample can be easily verified by hand.

The counterexample settled the universal statement, but it did not explain the object that settled it. Its topology, terminal demands, fractional path splits, and arc costs arrived as a tightly coupled package. The graph alone does not show why it leads to the conflict while a simpler topology cannot. The numerical values do not reveal which inequalities are essential or whether other parameter choices also work. The sparse cost vector does not establish whether the zero-cost arcs are merely convenient or can be imposed without loss of generality. Nor does one example show whether the topology is exceptional, whether the counterexample belongs to a broader family, or which features should be transferred to a different network. A single instance is enough to disprove a conjecture; understanding why it works requires a more well-developed theory.

More questions extend beyond the provenance of one counterexample, and they concern what leads to the success of AI and how an AI-generated mathematical object can be turned into general knowledge. For instance, public responses to Rybin's announcement have already asked whether a deterministic procedure could have explored the search more efficiently, what was special about the returned topology, why such a small example had not been found earlier, and what construction ideas ChatGPT had used \citep{RybinLinkedIn2026}.

To understand both the counterexample and, equally importantly, what guided the AI toward it, we first work backward from Rybin's example only far enough to identify a simple starting point, and then ask whether the remaining construction can be recovered through a forward search with as little blind enumeration as possible. Our analysis shows that it can. The pipeline begins with an arborescence, on which every terminal has a unique source--terminal path and failure is impossible. It then introduces terminal-entering shortcut arcs incrementally, creating interacting route choices while keeping the complete graph-induced path set explicit. Finally, it translates the resulting shared-arc load constraints into selection geometry. The feasible binary route choices form a finite set, whose convex hull defines the \emph{conflict polytope}. A topology can support a counterexample only if the fractional route-share vector can lie outside this polytope; a separating facet then guides the construction of the demands, fractional path flows, and arc costs. Starting from the arborescence underlying Rybin's network, this pipeline singles out Rybin's topology by adding three terminal-entering shortcut arcs. We prove that fewer than three shortcuts cannot support a counterexample and that, among the three-shortcut ancestor augmentations considered, only Rybin's placement can do so. Once the topology is fixed, however, the numerical parameters are not unique. A simple system of inequalities gives necessary and sufficient conditions for the demands and fractional path flows to support a counterexample, after which suitable separating costs can be constructed.

This characterization also explains why a counterexample remained difficult to discover through blind enumeration. An ex ante search must choose the topology, demands, fractional flow, and costs jointly. The topology space is already enormous, and each candidate graph induces its own collection of source--terminal paths and unsplittable routings. Even after fixing Rybin's topology and restricting the cost search, our volume calculation shows that the counterexample-supporting region occupies only about one part in $1.2$ million of the normalized parameter space.

Comparing our pipeline with Rybin's public interaction further clarifies how the AI-assisted search succeeded. In the topology dimension, the search did not follow a comparable forward construction. Indeed, it used broad computational exploration to eliminate thousands of candidates and extracted several partial structural restrictions from these failures, but the public record does not show how those restrictions directly motivated the final seven-vertex graph. After the topology appeared, the model simplified the remaining construction by imposing symmetry between two endpoint terminals and derived sufficient conditions for a counterexample. These conditions coincide with our complete characterization restricted to that symmetric subspace. Rybin's interaction can therefore be interpreted as lying between blind enumeration and a full theory-guided construction.

Finally, we test whether the pipeline has forward value by supplying ChatGPT with the established structural theory and a concise statement of the resulting search logic, and then asking it to continue the construction. On Rybin's topology, ChatGPT generated six additional integer counterexamples that are not uniform scalings of the original instance. We next examined whether the pipeline could move beyond the seven-vertex family by allowing both the initial arborescence and the shortcut placement to change. These searches produced counterexamples on genuinely different topologies, with separation driven by higher-order conflict-polytope facets rather than Rybin's triangle. Finally, to determine whether these successes depended on a few specially chosen starting trees, we removed the topology-novelty requirement and required ChatGPT to select and fix each initial arborescence before adding shortcuts and choosing the numerical parameters. Among arborescences with three to five terminals and at most twelve vertices, the search found exactly verified counterexamples starting from 94 distinct rooted arborescence shapes, showing that the pipeline is broadly transferable across starting arborescences. Overall, these experiments reveal AI's strong capacity for mathematical generalization.

The remainder of the paper is organized as follows. Section~\ref{sec:model} formulates the problem, reviews the relevant positive results, and discusses Rybin's counterexample. Sections~\ref{sec:preliminary} and \ref{sec:construction} develop the topology, cost, flow, and demand characterizations. Section~\ref{sec:human-barriers} explains the limits of brute-force search, distills the forward pipeline, and compares it with Rybin's public interaction. Section~\ref{sec:ai-search} tests the pipeline on the original topology, transfers it to new topologies, and maps its breadth across a broad catalog of starting arborescences. Section~\ref{sec:conclusion} concludes.

\section{Problem formulation and preliminaries}\label{sec:model}

Consider a finite directed graph $\gG=(\gV,\gA)$, where $\gV$ and $\gA$ denote the sets of vertices and arcs, respectively. We study a single-source routing problem in which one source serves several terminals. Denote the source by $s\in\gV$, and let $\gT=\{t_1,\ldots,t_k\}\subseteq\gV\setminus\{s\}$ be the set of $k\geq1$ terminals. Each terminal $t_i$ is associated with a positive demand $d_i$ to be routed from $s$ to $t_i$. We assume that every terminal is reachable from $s$. For each $i=1,\ldots,k$, let $\gP_i$ denote the set of directed $s$--$t_i$ paths. For each path $p\in\gP_i$, let
$\gA_{i,p}\subseteq\gA$ denote the set of arcs contained in $p$.
For every $i=1,\ldots,k$, $p\in\gP_i$, and $a\in\gA$, define the
arc--path incidence indicator $\delta_{i,p}^{a} = 1$ if $a\in\gA_{i,p}$ and $\delta_{i,p}^{a} = 0$ otherwise.

To specify the feasible routing plans, we must determine whether the demand associated with each terminal may be divided among several paths. We therefore distinguish between fractional flows, which permit such division, and unsplittable flows, which require the entire demand of each terminal to follow a single path.

\begin{definition}[Fractional and unsplittable flows]\label{def:flows}
A \emph{fractional flow} is a collection of nonnegative path flows
$\vf=(\evf_{i,p})_{i=1,\ldots,k;\,p\in\gP_i}$ satisfying
\begin{equation}\label{eq:demand-conservation}
\sum_{p\in\gP_i}\evf_{i,p}=d_i,
\quad i=1,\ldots,k.
\end{equation}
A fractional flow $\vf$ is \emph{unsplittable} if, for each
$i=1,\ldots,k$, there exists a path $p_i\in\gP_i$ such that
$\evf_{i,p_i}=d_i$ and all other path flows associated with
terminal $t_i$ are zero.
\end{definition}

A path-flow vector $\vf$ induces an arc-flow vector
$\vx=(\evx_a)_{a\in\gA}$, which satisfies
\begin{equation}\label{eq:arc-flow}
\evx_a
=
\sum_{i=1}^k\sum_{p\in\gP_i}
\delta_{i,p}^{a}\cdot\evf_{i,p}.
\end{equation}
Suppose that each arc $a\in\gA$ has a fixed nonnegative unit cost
$\evl_a$. The total routing cost associated with $\vf$ can then be computed as
\begin{equation}\label{eq:routing-cost}
c(\vf)
=
\sum_{a\in\gA}\evl_a\cdot\evx_a
=
\sum_{a\in\gA}\evl_a \cdot
\sum_{i=1}^k\sum_{p\in\gP_i}
\delta_{i,p}^{a}\cdot\evf_{i,p}.
\end{equation}

The literature has long examined how well the arc loads and total cost of a fractional flow can be preserved when it is converted into an unsplittable flow. Two foundational statements frame this question: the Dinitz--Garg--Goemans theorem (Section \ref{sec:theorem}) and Goemans' conjecture (Section \ref{sec:conjecture}), which will be introduced in the remainder of this section.

\subsection{The Dinitz--Garg--Goemans theorem}
\label{sec:theorem}

Converting a fractional flow into an unsplittable flow may increase
the load on some arcs because each terminal's demand must be assigned
to a single path. To measure the additive increase uniformly across arcs, let
\begin{equation}\label{eq:largest-demand}
D=\max_{i=1,\ldots,k}d_i
\end{equation}
denote the largest terminal demand. The following definition uses a parameter $\lambda\geq0$ to specify the additive increase relative to $D$.

\begin{definition}[$\lambda$-admissibility]
\label{def:lambda-admissibility}
Let $\vf$ be a fractional flow with induced arc-flow vector
$\vx=(\evx_a)_{a\in\gA}$. For any $\lambda\geq0$, an unsplittable
flow $\vf^{\mathrm{u}}$ with induced arc-flow vector
$\vx^{\mathrm{u}}=(\evx_a^{\mathrm{u}})_{a\in\gA}$ is
\emph{$\lambda$-admissible relative to $\vf$} if
\begin{equation}\label{eq:lambda-admissible}
\evx_a^{\mathrm{u}} \leq \evx_a+\lambda\cdot D, \quad a\in\gA.
\end{equation}
\end{definition}

Here, $\vf$ is referred to as the \emph{reference flow}, and $\evx_a$ is its
\emph{reference load} on arc $a$. Thus, $\lambda=0$ permits no increase
in any arc load, whereas $\lambda=1$ permits an increase of at most the
largest demand. The Dinitz--Garg--Goemans theorem establishes that
$\lambda=1$ always suffices.

\begin{theorem}[Dinitz--Garg--Goemans theorem]\label{thm:dgg}
For every fractional flow $\vf$, there exists an unsplittable flow
$\vf^{\mathrm{u}}$ that is $1$-admissible relative to $\vf$.
Equivalently,
\begin{equation}\label{eq:dgg-bound}
\evx_a^{\mathrm{u}}\leq\evx_a+D,
\quad a\in\gA.
\end{equation}
\end{theorem}

The theorem shows that indivisibility can always be imposed while limiting the additional flow on every arc to at most the largest demand. It does not, however, guarantee that the resulting unsplittable flow has no greater cost than the reference flow.

\subsection{Goemans' conjecture}
\label{sec:conjecture}

Goemans' conjecture then asks whether the same arc-flow guarantee can always be achieved without increasing total routing cost.

\begin{conjecture}[Goemans' conjecture]\label{conj:goemans}
Let $\vf$ be any fractional flow. For every collection of fixed nonnegative unit arc costs $(\evl_a)_{a\in\gA}$, there exists an unsplittable flow $\vf^{\mathrm{u}}$ that is $1$-admissible relative to $\vf$ and satisfies
$c(\vf^{\mathrm{u}})\leq c(\vf).$
\end{conjecture}

For a fixed nonnegative unit arc-cost vector and a reference flow $\vf$, define the minimum cost attainable under an additive load allowance $\lambda \cdot D$ as
\begin{equation}\label{eq:Cint-def}
c_{\lambda}^{\mathrm{u}}(\vf) = \min\left\{
c(\vf^{\mathrm{u}}):
\vf^{\mathrm{u}}
\text{ is an unsplittable flow that is $\lambda$-admissible relative to $\vf$}
\right\}.
\end{equation}
If no such unsplittable flow exists, we set $c_{\lambda}^{\mathrm{u}}(\vf)=+\infty$.
The Dinitz--Garg--Goemans theorem guarantees that $c_1^{\mathrm{u}}(\vf)$ is always finite, while Goemans' conjecture asserts that $c_1^{\mathrm{u}}(\vf)\leq c(\vf)$ for every fractional flow $\vf$.

\subsubsection{Positive results}
\label{sec:positive-results}

Before the recent counterexample, several studies had established Goemans' conjecture under additional structures. These structures fall into two classes.

The first restricts the terminal demands. \citet{Skutella2002}, as an early example, proved Goemans' conjecture when the terminal demands are successively divisible, meaning that they can be ordered as $d_1\geq\cdots\geq d_k>0$ with $d_i/d_{i+1}$ a positive integer for each $i=1,\ldots,k-1$. For the same demand structure, \citet{MorellSkutella2022} additionally bounded decreases in arc flows, showing that a cost-nonincreasing unsplittable flow can be chosen such that
\begin{equation}
\evx_a-D \leq \evx_a^{\mathrm{u}} \leq \evx_a+D, \quad a\in\gA.
\end{equation}

The second restricts the digraph on which the flows are routed. \citet{AlmoghrabiSkutellaWarode2026}, for example, proved the conjecture for series--parallel digraphs. Such a digraph has a designated start vertex and end vertex and is constructed recursively from a single directed arc. Two smaller series--parallel digraphs can be combined in series by identifying the end of the first with the start of the second, or in parallel by identifying their start vertices and, separately, their end vertices. \citet{TraubVargasKochZenklusen2026} considered the broader class of planar acyclic digraphs, which can be drawn without arc crossings and contain no directed cycles. They proved a relaxed cost-preserving guarantee: an unsplittable flow with no greater cost always exists if the flow on each arc is allowed to differ from the corresponding fractional flow by at most $2D$. Goemans' conjecture requires the sharper allowance $D$, so this result does not prove the conjecture for all planar acyclic digraphs, but it shows that doubling the allowance is sufficient.

\subsubsection{Rybin's counterexample}

Although the positive results above made the conjecture appear plausible, Rybin's seven-vertex instance in Figure~\ref{fig:rybin-certificate} disproves it. The terminal demands are $(d_1,d_2,d_3)=(15,10,15)$, so the largest demand is $D=15$. Each terminal has exactly two $s$--$t_i$ paths: a path $r_i\in\gP_i$ containing a positive-cost arc and a path $z_i\in\gP_i$ consisting entirely of zero-cost arcs:
\begin{alignat*}{2}
r_1&=(s,t_1),   &\quad z_1&=(s,u,v,t_1),\\
r_2&=(s,t_2),   &\quad z_2&=(s,u,v,w,t_2),\\
r_3&=(s,u,t_3), &\quad z_3&=(s,u,v,w,t_3).
\end{alignat*}
The only positive-cost arcs are $(s, t_1)$, $(s, t_2)$, and $(u, t_3)$, whose unit costs are $2$, $3$, and $2$, respectively.

Consider the fractional reference flow that assigns
\begin{equation}\label{eq:rybin-path-flow}
(\evf_{1,r_1},\evf_{1,z_1})=(10,5),\quad
(\evf_{2,r_2},\evf_{2,z_2})=(6,4), \quad
(\evf_{3,r_3},\evf_{3,z_3})=(10,5).
\end{equation}
The corresponding arc flows are
\begin{equation}\label{eq:rybin-arc-flow}
(
\evx_{(s,t_1)},
\evx_{(s,t_2)},
\evx_{(s,u)},
\evx_{(u,v)},
\evx_{(u,t_3)},
\evx_{(v,t_1)},
\evx_{(v,w)},
\evx_{(w,t_2)},
\evx_{(w,t_3)}
)
=
(10,6,24,14,10,5,9,4,5).
\end{equation}
Consequently, the total cost of the reference flow is $c(\vf)=2\cdot10+3\cdot6+2\cdot10=58.$ 

It can be checked that a $1$-admissible unsplittable flow can use at most one of $z_1$, $z_2$, and $z_3$. Selecting both $z_1$ and $z_2$ places $40$ units on arc $(s,u)$, exceeding the permitted load $24+D=39$. Similarly, selecting both $z_1$ and $z_3$ places at least $30>14+D=29$ units on $(u,v)$, while selecting both $z_2$ and $z_3$ places $25>9+D=24$ units on $(v,w)$. In the first comparison, terminal $t_3$ contributes its entire demand to $(s,u)$ because both $r_3$ and $z_3$ contain this arc. Thus, every $1$-admissible unsplittable flow must use at least two positive-cost paths. Routing the full demand of terminal $t_1$ through $r_1$ costs $2\cdot15=30$, routing the full demand of terminal $t_2$ through $r_2$ costs $3\cdot10=30$, and routing the full demand of terminal $t_3$ through $r_3$ costs $2\cdot15=30$. Accordingly, every $1$-admissible unsplittable flow has total cost at least $60$.
Therefore,
\begin{equation}\label{eq:rybin-gap}
c_1^{\mathrm{u}}(\vf)\geq60>58=c(\vf),
\end{equation}
which disproves Goemans' conjecture.

\section{From an arborescence to Rybin's topology}
\label{sec:preliminary}

Rybin's counterexample establishes that Goemans' conjecture can fail, yet the numerical certificate alone does not identify which features of its topology are essential. The seven-vertex network may be one of many nearby structures that support failure, or its arrangement of path alternatives may be exceptional. In this section, we will address this question by tracing the network back to a positive benchmark and then introducing path choice through the smallest controlled modifications of that benchmark.

In Rybin's network, three terminal-entering arcs with positive unit costs create a second path for each terminal. Removing these arcs preserves all seven vertices and the shared branching structure while leaving every terminal with a unique path from $s$. The resulting six-arc network is an arborescence, the directed analogue of a rooted tree. Given a directed graph $\gG=(\gV,\gA)$, define the indegree of each vertex $v\in\gV$ by
\begin{equation}
\deg^{-}(v)=\left|\left\{u\in\gV:(u,v)\in\gA\right\}\right|.
\end{equation}
An arborescence can then be formally defined below.

\begin{definition}[Arborescence]\label{def:arborescence}
A directed graph $\gG=(\gV,\gA)$ is an \emph{arborescence} rooted at $r\in\gV$ if
\begin{equation}
\deg^{-}(r)=0 \quad\text{and}\quad \deg^{-}(v)=1 \quad\text{for every }v\in\gV\setminus\{r\},
\end{equation}
and every vertex in $\gV$ is reachable from $r$ through a directed path.
\end{definition}

\begin{figure}[ht]
\centering
\includegraphics[width=0.4\textwidth]{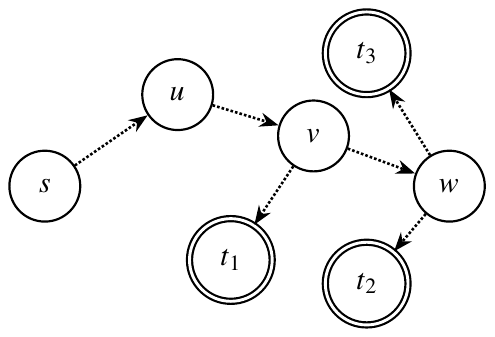}
\caption{The spanning arborescence embedded in Rybin's counterexample. It is obtained by removing the three terminal-entering arcs $(s,t_1)$, $(s,t_2)$, and $(u,t_3)$. Every terminal then has a unique path from $s$. In Rybin's numerical instance, all six displayed arcs have zero unit cost.}
\label{fig:base-arborescence}
\end{figure}

Removing $(s,t_1)$, $(s,t_2)$, and $(u,t_3)$ from Rybin's network leaves the graph $\gG^0=(\gV^0,\gA^0)$ as shown in Figure~\ref{fig:base-arborescence}, where
\begin{equation}\label{eq:base-arborescence}
\begin{aligned}
\gV^0&=\{s,u,v,w,t_1,t_2,t_3\},\\
\gA^0&=\{(s,u),(u,v),(v,w),(v,t_1),(w,t_2),(w,t_3)\}.
\end{aligned}
\end{equation}

It is straightforward to check that an arborescence can never support a counterexample because its unique-path structure leaves no demand to split fractionally.

\begin{proposition}
\label{prop:unique-path}
Let $\gG=(\gV,\gA)$ be an arborescence rooted at the source $s$ with fixed nonnegative unit arc costs $\vl \in\R_+^{|\gA|}$, and let $\gT\subseteq\gV\setminus\{s\}$ be any set of terminals with positive demands. Then, $
c_{0}^{\mathrm{u}}(\vf)=c(\vf)
$ for every fractional flow $\vf$ on $\gG$.

\end{proposition}

\begin{proof}
For each terminal $t_i$, let $p_i$ be the unique directed path from $s$ to $t_i$. Since $\gP_i=\{p_i\}$, the demand-conservation condition in \eqref{eq:demand-conservation} implies $\evf_{i,p_i}=d_i$. Hence, $\vf$ is itself an unsplittable flow and is $0$-admissible relative to itself. Moreover, every unsplittable flow must use the same unique path $p_i$ for each terminal, so $\vf$ is the only unsplittable flow. Therefore, $c_{0}^{\mathrm{u}}(\vf)=c(\vf)$.
\end{proof}

Proposition~\ref{prop:unique-path} establishes $\gG^0$ as a positive benchmark and implies that failure requires genuine path alternatives. In the remainder of this section, we introduce a general class of augmentations that create alternative paths through terminal-entering shortcut arcs. We first show that no augmentation with fewer than three shortcut arcs can support a counterexample (Section~\ref{sec:shortcut-augmentations}), and then examine three-shortcut augmentations and identify the topology that can make failure possible (Section~\ref{sec:shortcut-topology}).

\subsection{Ancestor-shortcut augmentations}
\label{sec:shortcut-augmentations}

To introduce path alternatives with minimal changes, we retain all vertices and arcs of $\gG^0$ and allow at most one additional arc entering each terminal. Each terminal has a unique parent in the arborescence. For a new arc $(q,t_i)$, we require $q$ to lie on the unique path from $s$ to the parent of $t_i$, excluding the parent itself. The new arc creates a second $s$--$t_i$ path by bypassing a nonempty final segment of the original path. We call such an arc an \emph{ancestor shortcut}.

The parent of $t_1$ is $v$, whereas the parent of each of $t_2$ and $t_3$ is $w$. The allowable shortcut arcs are therefore
\begin{equation}\label{eq:shortcut-sets}
\begin{aligned}
\gA_1^{\mathrm{sc}}=\{(s,t_1),(u,t_1)\}, \ \
\gA_2^{\mathrm{sc}}=\{(s,t_2),(u,t_2),(v,t_2)\}, \ \
\gA_3^{\mathrm{sc}}=\{(s,t_3),(u,t_3),(v,t_3)\}.
\end{aligned}
\end{equation}
We consider every augmented graph satisfying
\begin{equation}\label{eq:shortcut-augmentation}
\begin{aligned}
\gG=(\gV^0,\gA^0\cup\gA^{\mathrm{sc}}), \quad 
\gA^{\mathrm{sc}}\subseteq\bigcup_{i=1}^3\gA_i^{\mathrm{sc}}, \quad \left|\gA^{\mathrm{sc}}\cap\gA_i^{\mathrm{sc}}\right|\leq1 \quad(i=1,2,3).
\end{aligned}
\end{equation}
Thus, the original arborescence is retained, and each terminal receives either no shortcut or exactly one allowable shortcut. Rybin's topology corresponds to
\begin{equation}\label{eq:rybin-shortcut-set}
\gA^{\mathrm{sc}}=\{(s,t_1),(s,t_2),(u,t_3)\}.
\end{equation}

Rybin's topology gives every terminal one shortcut. We first ask whether fewer shortcuts can already support a counterexample.

\begin{proposition}
\label{prop:two-shortcuts}
Let $\gG^0=(\gV^0,\gA^0)$ be the arborescence defined in
Equation \eqref{eq:base-arborescence}, with source $s$ and terminal set $\gT=\{t_1,t_2,t_3\}$, and let
$\gG=(\gV^0,\gA)$ be an ancestor-shortcut augmentation satisfying
\eqref{eq:shortcut-augmentation}, where
$\gA=\gA^0\cup\gA^{\mathrm{sc}}$. Equip the terminals with a positive demand vector $\vd=(d_1,d_2,d_3)\in\R_{++}^3$ and the arcs with a fixed nonnegative unit arc-cost vector $\vl \in\R_+^{|\gA|}$.
If $\lvert\gA^{\mathrm{sc}}\rvert\leq2$, then
$c_1^{\mathrm{u}}(\vf)\leq c(\vf)$ for every fractional flow
$\vf$ on $\gG$.
\end{proposition}

\begin{proof} See Appendix~\ref{app:two-shortcuts}.
\end{proof}

Proposition~\ref{prop:two-shortcuts} eliminates every augmentation with fewer than three shortcut arcs. The only remaining question is where the three shortcuts can be placed.

\subsection{The unique three-shortcut topology}
\label{sec:shortcut-topology}

With one shortcut entering each terminal, Equation \eqref{eq:shortcut-sets} gives two possible placements for $t_1$ and three for each of $t_2$ and $t_3$. Hence, there are $2\cdot3^2=18$ labeled three-shortcut augmentations. Since $t_2$ and $t_3$ are symmetric in $\gG^0$, these augmentations reduce to $12$ topological classes up to exchanging the two terminals. The following theorem shows that the Rybin placement in \eqref{eq:rybin-shortcut-set} is the only class that can support a counterexample.

\begin{theorem}
\label{thm:shortcut-topology-uniqueness}
Let $\gG^0=(\gV^0,\gA^0)$ be the arborescence defined in Equation~\eqref{eq:base-arborescence}, with source $s$ and terminal set $\gT=\{t_1,t_2,t_3\}$, and let $\gG=(\gV^0,\gA^0\cup\gA^{\mathrm{sc}})$ satisfy Equation~\eqref{eq:shortcut-augmentation}. If there exist a positive demand vector $\vd\in\R_{++}^3$, a fractional reference flow $\vf$ on $\gG$, and a nonnegative unit arc-cost vector $\vl\in\R_+^{|\gA^0\cup\gA^{\mathrm{sc}}|}$ under which $c_1^{\mathrm{u}}(\vf)>c(\vf)$, then, up to exchanging terminals $t_2$ and $t_3$,
\begin{equation}\label{eq:unique-shortcut-arcs}
\gA^{\mathrm{sc}}
=
\{(s,t_1),(s,t_2),(u,t_3)\}.
\end{equation}
\end{theorem}

\begin{proof} See Appendix~\ref{app:topology-uniqueness}.
\end{proof}

Here, the unique placement is distinguished by a staggered sequence of path-choice interactions along the arborescence backbone. On the upstream arc $(s,u)$, the choices of terminals $t_1$ and $t_2$ alter the load; on the middle arc $(u,v)$, the choices of all three terminals alter the load; and on the downstream arc $(v,w)$, the choices of terminals $t_2$ and $t_3$ alter the load. Each pair of arborescence paths can therefore become jointly inadmissible on a different backbone arc, allowing the three potential pairwise conflicts to close a cycle. The theorem identifies the only topology in the augmentation class with this capability.

\section{When Rybin's topology gives a counterexample}\label{sec:construction}

The topology theorem identifies the only ancestor-shortcut augmentation that can support a counterexample. We now determine exactly which reference flows and unit arc costs make this topology do so. 
Fix the topology identified in Theorem~\ref{thm:shortcut-topology-uniqueness} and denote it by $\gG=(\gV,\gA)$, where
\begin{equation}\label{eq:arc-set}
\begin{aligned}
\gV&=\{s,u,v,w,t_1,t_2,t_3\},\\
\gA&=\{(s,t_1),(s,t_2),(s,u),(u,v),(u,t_3),(v,t_1),(v,w),(w,t_2),(w,t_3)\}.
\end{aligned}
\end{equation}
Let $s$ be the source and $\gT=\{t_1,t_2,t_3\}$ be the terminal set with positive demand vector $\vd=(d_1,d_2,d_3)\in\R_{++}^3$. Each terminal has exactly two source--terminal paths. We denote the shortcut path by $r_i$ and the path in the embedded arborescence by $z_i$:
\begin{equation}\label{eq:routes}
\begin{aligned}
r_1&=(s,t_1), & z_1&=(s,u,v,t_1),\\
r_2&=(s,t_2), & z_2&=(s,u,v,w,t_2),\\
r_3&=(s,u,t_3), & z_3&=(s,u,v,w,t_3).
\end{aligned}
\end{equation}
The arcs $(s,t_1)$, $(s,t_2)$, and $(u,t_3)$ are used only by $r_1$, $r_2$, and $r_3$, respectively, and by none of the arborescence paths. We call them the \emph{private shortcut arcs} and write
\begin{equation}\label{eq:private-shortcut-arcs}
\gA^{\mathrm{p}}
=
\{(s,t_1),(s,t_2),(u,t_3)\}.
\end{equation}

To simplify the notations, for any fractional reference flow $\vf$ on $\gG$, we use $\alpha_i$ to represent the fraction of terminal $t_i$'s demand routed through the arborescence path $z_i$, namely
\begin{equation}\label{eq:general-reference-shares}
\alpha_i
=
\frac{\evf_{i,z_i}}{d_i}
\in[0,1],
\quad i=1,2,3,
\end{equation}
and write $\valpha=(\alpha_1,\alpha_2,\alpha_3)$. Meanwhile, we represent an unsplittable routing choice by a selection vector $\vy\in\{0,1\}^3$, where $y_i=1$ means that terminal $t_i$ uses $z_i$ and $y_i=0$ means that it uses $r_i$. Let $\vf^{\mathrm{u}}(\vy)$ denote the corresponding unsplittable flow, and let $\evx_a^{\mathrm{u}}(\vy)$ denote its load on arc $a$. Define the set of $1$-admissible selections relative to $\vf$ by
\begin{equation}\label{eq:admissible-selection-set}
\gY(\vf)
=
\left\{
\vy\in\{0,1\}^3:
\vf^{\mathrm{u}}(\vy)
\text{ is $1$-admissible relative to $\vf$}
\right\}.
\end{equation}

\subsection{Reducing the cost search}
\label{sec:canonical-cost}

Rybin's counterexample assigns positive unit costs only to the three private shortcut arcs and zero unit cost to every other arc. The arborescence paths are therefore free, while each shortcut path incurs a positive cost on an arc used exclusively by that path. Let $\gL'$ denote the set of cost vectors with this canonical structure:
\begin{equation}\label{eq:canonical-cost-set}
\gL'
=
\left\{
\vl=(l_a)_{a\in\gA}\in\R_+^{|\gA|}:
l_a=0, \ \ 
a\in\gA\setminus\gA^{\mathrm{p}}
\right\}.
\end{equation}

When the unit arc-cost vector must be displayed explicitly, we write $c(\vf;\vl)$ for the cost of $\vf$ under $\vl$ and $c_{\lambda}^{\mathrm{u}}(\vf;\vl)$ for the minimum in Equation~\eqref{eq:Cint-def} under the same vector. For fixed terminal demands, a pair $(\vf,\vl)$ forms a counterexample when
$
c_1^{\mathrm{u}}(\vf;\vl)>c(\vf;\vl).
$
The next proposition shows that restricting the cost vector to $\gL'$ does not change whether a counterexample exists on this topology.

\begin{proposition}
\label{prop:canonical-cost}
Let $\gG=(\gV,\gA)$ be the topology defined in Equation~\eqref{eq:arc-set}, with source $s$, terminal set $\gT=\{t_1,t_2,t_3\}$, and positive demand vector $\vd=(d_1,d_2,d_3)\in\R_{++}^3$. For any fractional reference flow $\vf$ on $\gG$, if
$
c_1^{\mathrm{u}}(\vf;\vl)>c(\vf;\vl)
$
for some $\vl\in\R_+^{|\gA|}$, then the same strict inequality holds for some $\vl'\in\gL'$.
\end{proposition}

\begin{proof}
See Appendix~\ref{app:canonical-cost}.
\end{proof}

Proposition~\ref{prop:canonical-cost} shows that, for the purpose of finding counterexamples, it is sufficient to search over cost vectors in $\gL'$. Other nonnegative cost vectors may also generate counterexamples, but any reference flow that can be separated by some cost vector can also be separated by one within $\gL'$.

\subsection{Conditions for a counterexample}
\label{sec:triangle-characterization}

We now proceed to derive a complete characterization of the fractional reference flows on the seven-vertex topology that generate counterexamples.

We first parameterize an arbitrary fractional reference flow $\vf$. To simplify notation, define
\begin{equation}\label{eq:reference-path-flow}
a=\evf_{1,z_1},
\quad
b=\evf_{2,z_2},
\quad
c=\evf_{3,z_3}.
\end{equation}
Because each terminal has exactly two paths, the remaining path flows are $\evf_{1,r_1}=d_1-a$, $\evf_{2,r_2}=d_2-b$, and $\evf_{3,r_3}=d_3-c$. Flow feasibility gives $a\in[0,d_1]$, $b\in[0,d_2]$, and $c\in[0,d_3]$.

The following theorem gives a compact necessary-and-sufficient characterization of the existence of a counterexample within this parameterized family.

\begin{theorem}
\label{thm:complete-triangle}
Let $\gG=(\gV,\gA)$ be the topology defined in Equation~\eqref{eq:arc-set}, with source $s$, terminal set $\gT=\{t_1,t_2,t_3\}$, positive demand vector $\vd=(d_1,d_2,d_3)\in\R_{++}^3$, and $D=\max\{d_1,d_2,d_3\}$. For any fractional flow $\vf$ on $\gG$, let $a,b,c$ be defined by Equation~\eqref{eq:reference-path-flow}. There exists a nonnegative unit arc-cost vector $\vl\in\R_+^{|\gA|}$ such that
$
c_1^{\mathrm{u}}(\vf;\vl)>c(\vf;\vl)
$
if and only if
\begin{equation}\label{eq:complete-system}
\begin{cases}
d_1+d_2>a+b+D,\\
d_1+d_3>a+b+c+D,\\
d_2+d_3>b+c+D,\\
\displaystyle \frac{a}{d_1}+\frac{b}{d_2}+\frac{c}{d_3}>1.
\end{cases}
\end{equation}
\end{theorem}

\begin{proof}[Proof sketch]
Recall that a selection vector $\vy=(y_1,y_2,y_3)\in\{0,1\}^3$ assigns terminal $t_i$ to $z_i$ when $y_i=1$ and to $r_i$ when $y_i=0$. Call the pair $i$--$j$ \emph{blocked} when the selection using $z_i$ and $z_j$ and the remaining shortcut path is not $1$-admissible relative to $\vf$. Comparing the three pair selections with the reference loads on the shared backbone shows that the first three inequalities in Equation~\eqref{eq:complete-system} block the pairs $1$--$2$, $1$--$3$, and $2$--$3$ on $(s,u)$, $(u,v)$, and $(v,w)$, respectively. These three conflicts close the conflict graph into a triangle.

For necessity, suppose that a counterexample exists. By Proposition~\ref{prop:canonical-cost}, the separating cost vector can be chosen from $\gL'$. Recall that $\valpha=(a/d_1,b/d_2,c/d_3)$ records the fractions of the three demands routed along $z_1,z_2,z_3$, and let $K_i$ be the cost of routing the full demand $d_i$ along $r_i$. The reference and selected-flow costs are then $\sum_{i=1}^3K_i\cdot(1-\alpha_i)$ and $\sum_{i=1}^3K_i\cdot(1-y_i)$, respectively. Thus, if $\valpha$ were a convex combination of $1$-admissible selection vectors, the reference cost would be the same convex combination of their costs, and at least one would be no greater than the reference cost. Consequently, a counterexample can exist only if
\begin{equation}\label{eq:outside-conflict-polytope}
\valpha\notin\gC(\vf),
\quad \text{where}~
\gC(\vf):=\conv(\gY(\vf)).
\end{equation}
Here, $\conv(\gY(\vf))$ is the set of all convex combinations of $1$-admissible selection vectors. It can be shown that every conflict pattern with fewer than three blocked pairs has $\valpha\in\gC(\vf)$. Hence, all three pairs must be blocked, which is equivalent to the first three inequalities in Equation~\eqref{eq:complete-system}. In this case,
\begin{equation}\label{eq:triangle-selection-set}
\gY(\vf)=\gY_1=\{(0,0,0),(1,0,0),(0,1,0),(0,0,1)\},
\end{equation}
and $\gC(\vf)=\{\vy\in\R_+^3:y_1+y_2+y_3\leq1\}$. Condition~\eqref{eq:outside-conflict-polytope} therefore becomes $\alpha_1+\alpha_2+\alpha_3>1$, which is the fourth inequality in Equation~\eqref{eq:complete-system}. For sufficiency, suppose that the four inequalities in Equation~\eqref{eq:complete-system} hold. The first three give $\gY(\vf)=\gY_1$. Assign the same full-demand cost $K_0>0$ to each shortcut path and zero cost to every arborescence path. The fourth inequality then gives
\begin{equation}\label{eq:equal-cost-gap}
c(\vf;\vl') = K_0\cdot(3-\alpha_1-\alpha_2-\alpha_3) < 2K_0 = c_1^{\mathrm{u}}(\vf;\vl').
\end{equation}
Hence, a counterexample exists. The complete proof is given in Appendix~\ref{app:complete-triangle}.
\end{proof}

Here, the characterization separates the counterexample mechanism into a physical component and a geometric component. The first three inequalities make each pair of arborescence paths jointly inadmissible, closing the conflict graph into a triangle. The fourth places the fractional share vector beyond the corresponding triangle facet. Thus, the seven-vertex topology generates counterexamples by realizing the smallest odd cycle of pairwise conflicts and placing the reference flow on the separating side of the facet created by that cycle.

Theorem~\ref{thm:complete-triangle} determines whether a counterexample exists when both these demands and the fractional reference flow, represented by $(a,b,c)$, are specified. We now eliminate $(a,b,c)$ and ask which demand vectors $(d_1,d_2,d_3)\in\R_{++}^3$ admit at least one fractional reference flow satisfying the four conditions in Equation~\eqref{eq:complete-system}. Let $D\geq M\geq S>0$ denote the three entries of $(d_1,d_2,d_3)$ arranged in nonincreasing order. Define
\begin{equation}\label{eq:rho-sigma}
\rho=\frac{S}{D}
\quad \text{and} \quad
\sigma=\frac{M}{D},
\end{equation}
so that $0<\rho\leq\sigma\leq1$. Because the endpoint terminals $t_1$ and $t_3$ are symmetric, the relevant assignment question is which of the three ordered demands should be placed at the structurally distinct central terminal $t_2$. The following theorem characterizes exactly which demand magnitudes can generate a counterexample in the triangle family and how they must be assigned to the three terminal roles.

\begin{theorem}
\label{thm:demand-region}
Let $\gG=(\gV,\gA)$ be the topology defined in Equation~\eqref{eq:arc-set}, with source $s$, terminal set $\gT=\{t_1,t_2,t_3\}$, and positive demand vector $\vd=(d_1,d_2,d_3)\in\R_{++}^3$. Let $D\geq M\geq S$ be the entries of $\vd$ arranged in nonincreasing order. There exist a fractional flow $\vf$ on $\gG$ and a nonnegative unit arc-cost vector $\vl\in\R_+^{|\gA|}$ such that
$
c_1^{\mathrm{u}}(\vf;\vl)>c(\vf;\vl)
$
if and only if the demand assignment satisfies $(d_1,d_2,d_3)=(D,S,M)$ or $(d_1,d_2,d_3)=(M,S,D)$,
and
\begin{equation}\label{eq:demand-region}
\frac{1}{2}<\rho<1
\quad \text{and} \quad
\sigma>\frac{\rho^2}{3\rho-\rho^2-1}.
\end{equation}
\end{theorem}

\begin{proof}
See Appendix~\ref{app:demand-region-proof}.
\end{proof}

Theorem~\ref{thm:demand-region} projects the flow-level characterization in Theorem~\ref{thm:complete-triangle} onto the demand space. It imposes both a role condition and a balance condition: the smallest demand must be assigned to the central terminal $t_2$, and all three demands must remain sufficiently close in size.

The proof also yields an explicit construction. It suffices to state the construction for the assignment $(d_1,d_2,d_3)=(D,S,M)$; the construction for $(M,S,D)$ follows by exchanging $t_1$ and $t_3$, and hence $a$ and $c$.
Define
\begin{equation}\label{eq:Phi-star}
\Phi^*(\rho,\sigma)
=
4-\rho-\frac{1}{\rho}-\frac{\rho}{\sigma}.
\end{equation}
Under Equation~\eqref{eq:demand-region}, $\Phi^*(\rho,\sigma)>1$. Choose any $\eta$ such that
$1 / \Phi^*(\rho,\sigma)<\eta<1$,
and set
\begin{equation}\label{eq:explicit-abc}
a=\eta\cdot(D-S),
\quad
b=\eta\cdot(2S-D),
\quad
c=\eta\cdot(M-S).
\end{equation}
Then, $(a,b,c)$ satisfies Equation~\eqref{eq:complete-system}. Assigning equal positive full-demand costs to the three shortcut paths and zero cost to every arborescence path therefore produces a counterexample.

\section{Why the counterexample was difficult to discover}
\label{sec:human-barriers}

Rybin's counterexample is so compact that brute-force discovery may appear entirely plausible. The network has only seven vertices and both the demands and the positive unit arc costs are small integers. Once the instance is given, only $2^3=8$ unsplittable routings need to be checked. These features naturally raise the question of whether an exhaustive search over small networks and integer parameters should have found the counterexample much earlier.

\subsection{The limits of brute-force search}

The difficulty is that an ex ante brute-force search must choose the topology, the terminal demands, the arc-cost vectors, and the fractional reference flows jointly.

The topology space alone is enormous. Even on seven labeled vertices, there are $2^{42}$ possible simple directed arc sets; fixing a topological ordering and allowing only forward arcs still leaves $2^{21}$ possibilities. Our analysis avoids such direct enumeration by first fixing an arborescence and then adding terminal-entering arcs incrementally. This construction keeps the complete path set controllable and reduces the topology search to a small family of structured augmentations. Without this reduction, enumerating topologies would be impractical.

Even after fixing Rybin's topology, a brute-force search must still choose the demands, fractional path flows, and costs jointly. Recall that $D\geq M\geq S$ are the ordered demand magnitudes and that $\rho=S/D$ and $\sigma=M/D$. After normalizing all demands and path flows by $D$, one of the two counterexample-supporting role assignments is
$
(d_1,d_2,d_3)=(1,\rho,\sigma);
$
the other exchanges the two endpoint terminals. We continue to write
$(A,B,C)=(a,b,c)/D$ and
$\valpha=(A,B/\rho,C/\sigma)$. For fixed $(\rho,\sigma)$, the ambient flow region is the box
$
[0,1]\times[0,\rho]\times[0,\sigma],
$
whose volume is $\rho\cdot\sigma$. Let $\mathcal F(\rho,\sigma)$ denote the subset satisfying the four conditions in Theorem~\ref{thm:complete-triangle}.

Recall also that $K_i$ denotes the cost of routing the full demand $d_i$ along shortcut path $r_i$. Strict separation is homogeneous in the cost scale, so we normalize
$
K_1+K_2+K_3=1.
$
For every point in $\mathcal F(\rho,\sigma)$, the $1$-admissible selections are the zero vector and the three unit vectors. The reference cost is therefore
$
1-\sum_{i=1}^3\alpha_i\cdot K_i,
$
whereas the minimum $1$-admissible unsplittable cost is
$
1-\max_{i=1,2,3}K_i.
$
Hence, a normalized cost vector separates the reference flow if and only if
$$
\sum_{i=1}^3\alpha_i\cdot K_i>\max_{i=1,2,3}K_i.
$$
For fixed $\valpha$, let $q(\valpha)$ denote the relative area of the cost vectors satisfying this inequality within the normalized cost simplex
$
\{(K_1,K_2,K_3)\in\R_+^3:K_1+K_2+K_3=1\}.
$
A direct area calculation gives
$$
q(\valpha)
=
\frac{
(\alpha_1+\alpha_2+\alpha_3-1)^2
}{
\prod_{i=1}^3
(1-\alpha_1-\alpha_2-\alpha_3+3\alpha_i)
}.
$$

We can now measure the demand, role, flow, and cost restrictions jointly. Under ordinary volume in these normalized coordinates, the resulting ratio is
\begin{equation}\label{eq:joint-parameter-volume}
\frac{
2\displaystyle
\int_{1/2}^{1}
\int_{\rho^2/(3\rho-\rho^2-1)}^{1}
\int_{\mathcal F(\rho,\sigma)}
q\left(A,\frac{B}{\rho},\frac{C}{\sigma}\right)
\,\mathrm dA\,\mathrm dB\,\mathrm dC\,
\mathrm d\sigma\,\mathrm d\rho
}{
6\displaystyle
\int_0^1\int_\rho^1
\rho\sigma\,\mathrm d\sigma\,\mathrm d\rho
}
\approx
8.44\times10^{-7}.
\end{equation}
Here, the numerator integrates over the two counterexample-supporting role assignments and weights each feasible demand--flow point by the fraction $q(\valpha)$ of normalized cost vectors that separate it; the denominator covers all six role assignments and all fractional path flows in their corresponding boxes. Thus, even after the correct topology has been fixed and the cost search has been restricted to the three canonical shortcut costs, only about $0.0000844\%$ (about 1 in $1.2$ million) of the normalized demand--role--flow--cost space supports a counterexample.

\subsection{A forward construction pipeline}

Does this mean that a human researcher had no realistic route to the counterexample? Our results suggest otherwise. The structural results above yield a forward procedure for constructing and testing candidate counterexamples.
\vspace{2pt}
\begin{enumerate}
\item \textit{Start from an arborescence.}
Begin with an arborescence, on which failure is impossible. This benchmark identifies path choice as a necessary source of failure and provides a controlled starting point for enlarging the search space.

\item \textit{Introduce route choices.}
Add path alternatives incrementally while keeping the complete set of source--terminal paths explicitly describable. Candidate modifications should create interactions among terminal choices without generating uncontrolled merge--split combinations or hidden hybrid paths.

\item \textit{Translate the topology into selection geometry.}
For each candidate topology, derive the arc-load inequalities defining its $1$-admissible unsplittable selections and examine their convex hull. A candidate can support a counterexample only when its fractional route-share vector can lie outside this polytope. The topology search can therefore be organized around graph-realizable forbidden-selection patterns and the facets they induce.
\end{enumerate}
\vspace{2pt}

In our reconstruction of Rybin's example, the choice of the initial arborescence was obtained by working backward from the known counterexample. Yet, once this benchmark is fixed, the subsequent reasoning proceeds constructively, where each step is largely motivated by the preceding result and substantially narrows the remaining search space. It is easy to be wise after the event: the arborescence underlying Rybin's example seems to be simple enough that it could plausibly have been encountered in a systematic enumeration of small arborescences; many other small arborescences may also serve as viable starting points for counterexample constructions\footnote{The later experiments in Section~\ref{sec:systematic-tree-transfer} will confirm this possibility.}.

Nonetheless, the pipeline itself is a contribution of the present analysis, developed only after a valid counterexample had established that a successful endpoint existed. While it remained unknown whether the conjecture was true or false, failure on a few arborescences could easily have led researchers to abandon this line of attack rather than move systematically to the next candidate. Human discovery was therefore possible in principle, but it required confidence in a search direction whose viability had not yet been established.

\subsection{Rybin's search between brute force and theory}
\label{sec:rybin-between-search-and-theory}

Rybin's four-round conversation with ChatGPT required approximately 53, 89, 94, and 88 minutes of reasoning, totaling more than five hours. Based on the public record, the search was neither a pure brute-force enumeration nor a fully theory-guided construction of the kind developed in this paper.

The first round largely followed a gadget-based construction strategy: the model designed small local network modules, each intended to enforce a particular conflict among a few designated paths, and then attempted to assemble them into a complete directed graph. To evaluate each candidate, it formulated an exact separation test. For a fixed fractional arc-flow vector $\vx$, let $U(\vx)$ be the set of all $1$-admissible unsplittable arc-flow vectors. On an acyclic graph, a counterexample requires
$
\vx\notin\conv U(\vx).
$
The model reported testing 7,248 small layered instances and 378 larger randomly generated directed acyclic graphs. None produced a valid counterexample. The central difficulty arose when the local modules were connected, where shared arcs created additional source--terminal paths. Once these graph-induced hybrid paths and the resulting unsplittable routings were included, the apparent separations disappeared. Repeated failures during the second and third rounds began to yield partial structural theory. For example, the model established that a counterexample cannot have all terminal demands equal, regardless of the network topology. It also explained why its two-terminal anchor construction failed. These findings suggested, without proving, that a successful counterexample would require several interacting nonuniform demands and a genuinely global obstruction. Even after these reductions, however, no counterexample had emerged.

The public record does not reveal the exact step by which ChatGPT selected the final seven-vertex topology in the fourth round. One plausible interpretation is that repeated failures progressively narrowed the viable design space, leaving the topology in Figure~\ref{fig:rybin-certificate} among a much smaller set of plausible candidates. Once this topology appeared, however, the subsequent parameter construction became considerably more transparent. Earlier in the interaction, the model had established that a counterexample cannot have all terminal demands equal, regardless of the network topology. To retain the required demand heterogeneity while reducing the number of free parameters, the final response imposed symmetry between the two endpoint terminals:
$$
d_1=d_3=1,\quad d_2=\beta\leq1,
\quad
\alpha_1=\alpha_3=r,\quad \alpha_2=q.
$$
Under this parameterization, it derived the conditions
$$
\beta\cdot(1-q)>r,\quad
2r+\beta\cdot q<1,\quad
2r+q>1.
$$
The first condition simultaneously blocks the two pairs involving the central terminal, the second blocks the pair of endpoint terminals, and the third places the fractional route-share vector beyond the triangle facet. The transcript presents these inequalities as sufficient conditions for the symmetric construction. Theorem~\ref{thm:complete-triangle} shows that, within this symmetric subspace, they are in fact necessary and sufficient: the four conditions in the theorem reduce exactly to the three inequalities above. Thus, although the transcript does not establish completeness, the conditions it states coincide with an exact symmetric slice of our characterization. The origin of the sparse cost vector is less clear. The final response directly assigns zero cost to all arborescence arcs and positive costs only to the three private shortcut arcs, chosen so that the three shortcut paths have equal full-demand costs. The public record does not show the rationale. Proposition~\ref{prop:canonical-cost}, as discussed earlier, fills this gap.

Overall, Rybin's ChatGPT search is best understood as a combination of broad computation and partial theory guidance. In the topology dimension, it remained much closer to brute-force exploration. Once the topology appeared, however, the search became more analytical. The model imposed a symmetric parameterization and stated sufficient conditions encoding the three pair blockers and the triangle-facet violation, although the rigor of this derivation cannot be audited from the transcript. The striking capability lies in the rapid transition between these two modes, where the model sustained a large volume of unsuccessful exploration, retained useful restrictions from those failures, and then compressed the successful topology into a low-dimensional symbolic construction. A human researcher could in principle perform each of these steps, but maintaining this breadth and persistence is much more challenging.

\section{Theory-guided AI search}
\label{sec:ai-search}

With the forward construction pipeline now in place, we next test whether ChatGPT can use it to generate further counterexamples and extend its logic beyond the original example. We supplied the structural analysis and the pipeline developed above to OpenAI's ChatGPT, running GPT-5.6 Pro with access to its built-in Python execution environment. We conduct three experiments. First, we keep the seven-vertex topology fixed and ask ChatGPT to construct further integer counterexamples  (Section~\ref{sec:ai-same-topology}). Second, we allow both the initial arborescence and the shortcut placement to change, and examine whether the pipeline can produce counterexamples on different networks (Section~\ref{sec:transfer-pipeline}). Finally, we drop the requirement that the resulting topology differ from Rybin's and ask how broadly the pipeline can succeed across small initial arborescences (Section~\ref{sec:systematic-tree-transfer}).

\subsection{Further counterexamples on Rybin's topology}
\label{sec:ai-same-topology}

In Rybin's network, Theorems~\ref{thm:complete-triangle} and \ref{thm:demand-region} provide exact criteria for counterexamples. We supplied Sections~\ref{sec:preliminary}--\ref{sec:construction} along with the general background to ChatGPT and used the following prompt:
\begin{quote}
\emph{``Read the attached note carefully and use its theoretical results to continue the search for counterexamples on the same seven-vertex topology. Find several genuinely new integer instances that are not uniform scalings of Rybin's example. For each instance, give the demands, fractional path flows, and nonnegative integer arc costs, together with a complete exact verification of the additive-load constraints and the strict cost gap."}
\end{quote}
After approximately 25 minutes of reasoning, ChatGPT returned six counterexamples. More importantly, it extracted from Theorem~\ref{thm:complete-triangle} a simple discrete construction that was not stated explicitly earlier. Choose the arborescence-path flows as
\begin{equation}\label{eq:one-unit-slack-family}
a=d_1-d_2,
\quad
b=2d_2-D-1,
\quad
c=d_3-d_2.
\end{equation}
Whenever these quantities are positive and their total demand share exceeds one, the first three conditions in Equation~\eqref{eq:complete-system} hold with exactly one unit of slack:
\begin{equation}\label{eq:one-unit-blockers}
a+b+D=d_1+d_2-1,
\quad
a+b+c+D=d_1+d_3-1,
\quad
b+c+D=d_2+d_3-1.
\end{equation}
Thus, the three pair selections are automatically blocked. The remaining search is reduced to finding positive integer demands satisfying
\begin{equation}\label{eq:one-unit-share-condition}
d_1>d_2,
\quad
d_3>d_2,
\quad
2d_2>D+1,
\quad
\frac{d_1-d_2}{d_1}
+
\frac{2d_2-D-1}{d_2}
+
\frac{d_3-d_2}{d_3}
>1.
\end{equation}
This converts the search over fractional path flows into a search over three integer demands.

The response then assigned integer unit costs only to the three private shortcut arcs, as justified by Proposition~\ref{prop:canonical-cost}. Table~\ref{tab:ai-integer-instances} reports the six resulting instances; their validity can be easily verified. We write
$$
(\ell_1,\ell_2,\ell_3)
=
\left(
l_{(s,t_1)},
l_{(s,t_2)},
l_{(u,t_3)}
\right),
$$
and set every other unit arc cost to zero. None of these instances is a uniform scaling of Rybin's example. In particular, their demand vectors do not have the ratio $3:2:3$. Instances I5 and I6 also show that the construction does not require equal endpoint demands.

\begin{table}[ht]
\centering
\caption{Integer counterexamples generated by the theory-guided AI search on the seven-vertex topology.}
\label{tab:ai-integer-instances}
\small
\setlength{\tabcolsep}{5pt}
\begin{tabular}{@{}lccccc@{}}
\toprule
Instance
& $(d_1,d_2,d_3)$
& $(a,b,c)$
& $(\ell_1,\ell_2,\ell_3)$
& $c(\vf;\vl)$
& $c_1^{\mathrm{u}}(\vf;\vl)$\\
\midrule
I1 & $(9,7,9)$   & $(2,4,2)$ & $(7,9,7)$ & $125$ & $126$\\
I2 & $(10,7,10)$ & $(3,3,3)$ & $(5,7,5)$ & $98$  & $99$\\
I3 & $(10,8,10)$ & $(2,5,2)$ & $(4,5,4)$ & $79$  & $80$\\
I4 & $(16,12,16)$& $(4,7,4)$ & $(3,4,3)$ & $92$  & $96$\\
I5 & $(15,12,16)$& $(3,7,4)$ & $(3,4,3)$ & $92$  & $93$\\
I6 & $(17,12,18)$& $(5,5,6)$ & $(2,3,2)$ & $69$  & $70$\\
\bottomrule
\end{tabular}
\end{table}

This experiment shows that the necessary-and-sufficient conditions developed in Section~\ref{sec:triangle-characterization} are directly useful for construction. Given these conditions, the AI reduced the remaining search to a simple integer problem, derived a new discrete subfamily, and generated several exactly verifiable counterexamples in a single run.

\subsection{Transferring the pipeline to new topologies}
\label{sec:transfer-pipeline}

We next ask whether the search pipeline can be transferred beyond the seven-vertex topology. Specifically, can the same pipeline --- starting from an arborescence, introducing controlled route choices, and analyzing the resulting selection geometry --- lead to counterexamples on genuinely different networks? A successful transfer would show that the proposed pipeline captures a reusable search logic whose value extends beyond explaining a single discovery.

\begin{figure}[ht]
\centering
\includegraphics[width=0.4\textwidth]{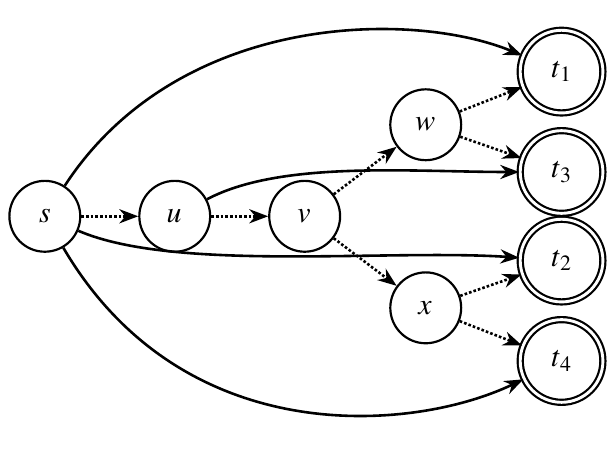}
\caption{Four-terminal topology produced by the first search. Dotted arcs form the zero-cost forked arborescence, whereas solid arcs are positive-cost terminal-entering shortcuts. Every arc carries positive reference flow, but deleting $t_4$ and suppressing the resulting subdivision vertex recovers Rybin's topology up to terminal relabeling.}
\label{fig:first-search-topology}
\end{figure}

We first used the following prompt:
\begin{quote}
\emph{``Read the attached note carefully and use its structural results to search for a complete counterexample on a genuinely different topology. The new network must not be obtainable from Rybin's seven-vertex topology merely by relabeling vertices, subdividing arcs, or adding unused or redundant arcs. You may generalize the odd-cycle mechanism identified in the note or discover a different conflict-polytope obstruction. Give the graph, terminal demands, fractional path flows, and nonnegative arc costs, and provide an exact verification that accounts for all source--terminal paths and proves that every $1$-admissible unsplittable flow has strictly greater cost than the fractional reference flow."}
\end{quote}
After approximately 75 minutes of reasoning, ChatGPT returned a valid four-terminal counterexample on the forked arborescence shown in Figure~\ref{fig:first-search-topology}. The terminal-demand vector was
$
(d_1,d_2,d_3,d_4)=(7,9,9,9).
$
The complete path-flow specification was
\begin{equation}\label{eq:first-transfer-path-flows}
\begin{aligned}
r_1&=(s,t_1),&
z_1&=(s,u,v,w,t_1),&
(f_{1,r_1},f_{1,z_1})&=(3,4),\\
r_2&=(s,t_2),&
z_2&=(s,u,v,x,t_2),&
(f_{2,r_2},f_{2,z_2})&=(1,8),\\
r_3&=(s,u,t_3),&
z_3&=(s,u,v,w,t_3),&
(f_{3,r_3},f_{3,z_3})&=(7,2),\\
r_4&=(s,t_4),&
z_4&=(s,u,v,x,t_4),&
(f_{4,r_4},f_{4,z_4})&=(6,3).
\end{aligned}
\end{equation}
Positive unit costs were assigned only to the four shortcut arcs:
$
l_{(s,t_1)}=9,
$
$
l_{(s,t_2)}=7,
$
$
l_{(u,t_3)}=7,
$
and
$
l_{(s,t_4)}=7.
$
Thus, routing any terminal's full demand along its shortcut path costs $63$, and
$
c_1^{\mathrm{u}}(\vf)=126>125=c(\vf).
$
The answer also moved beyond the triangle mechanism: its admissible-selection polytope had the higher-order facet
$
y_1+y_2+y_3+y_4\leq2.
$

The first answer satisfied the literal activity requirements. Every arc carried positive reference flow, every terminal used both of its paths, and removing any terminal destroyed the displayed certificate. It nevertheless retained Rybin's topology as a directed topological core. Deleting terminal $t_4$ and its two incident arcs, suppressing the resulting indegree-one/outdegree-one nonterminal $x$, and exchanging the labels of $t_1$ and $t_2$ recovers the seven-vertex topology. Thus, the phrase ``genuinely different'' had not excluded an active extension of the known graph.

We therefore opened a fresh conversation and strengthened the prompt:
\begin{quote}
\itshape
``Read the attached note carefully and search for a complete counterexample on a topology that does not contain Rybin's seven-vertex network as a directed topological minor. Concretely, for no choice of three terminals may Rybin's topology be obtained, up to relabeling that preserves the source and terminal roles, by deleting vertices or arcs and then suppressing nonterminal vertices with indegree one and outdegree one in the remaining graph. Do not return an active extension of Rybin's network.
The new counterexample may use any number of terminals and may arise from an odd-cycle obstruction or a different conflict-polytope facet. Give the graph, terminal demands, fractional path flows, and nonnegative arc costs. Enumerate all source--terminal paths, verify exactly that every $1$-admissible unsplittable flow has strictly greater cost than the fractional reference flow, and provide a graph-theoretic proof that the topology satisfies the directed-topological-minor exclusion above."
\end{quote}

\begin{figure}[ht]
\centering
\includegraphics[width=0.4\textwidth]{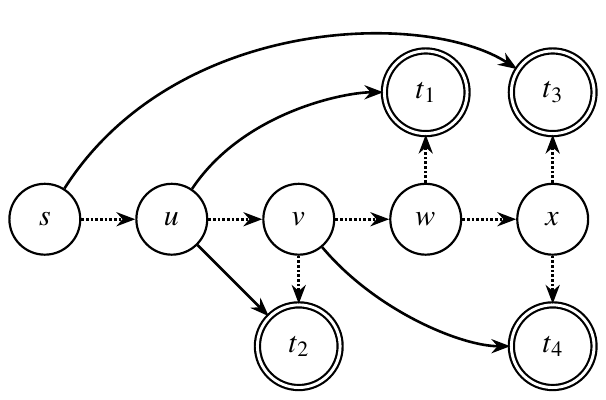}
\caption{Four-terminal topology produced by the stricter theory-guided search. Dotted arcs form the zero-cost arborescence, whereas solid arcs are positive-cost terminal-entering shortcuts. The source has outdegree two, which excludes Rybin's topology as a source-preserving directed topological minor.}
\label{fig:minorfree-topology}
\end{figure}

After approximately 50 minutes of reasoning, ChatGPT returned the nine-vertex, four-terminal topology shown in Figure~\ref{fig:minorfree-topology}. Its zero-cost arborescence consists of the nonterminal chain $s\to u\to v\to w\to x$ and the terminal-entering arcs $(w,t_1)$, $(v,t_2)$, $(x,t_3)$, and $(x,t_4)$. The four positive-cost shortcuts are $(u,t_1)$, $(u,t_2)$, $(s,t_3)$, and $(v,t_4)$.  Here, the source has outdegree two, whereas the source of Rybin's topology has outdegree three. Deleting vertices or arcs cannot increase source outdegree, and suppressing a nonterminal vertex with indegree one and outdegree one cannot increase the outdegree of its predecessor. The new topology therefore does not contain Rybin's network as a source-preserving directed topological minor.

The terminal-demand vector was $(d_1,d_2,d_3,d_4)=(13,12,10,13)$. The complete path-flow specification was
\begin{equation}\label{eq:minorfree-path-flows}
\begin{aligned}
r_1&=(s,u,t_1),&
z_1&=(s,u,v,w,t_1),&
(f_{1,r_1},f_{1,z_1})&=(10,3),\\
r_2&=(s,u,t_2),&
z_2&=(s,u,v,t_2),&
(f_{2,r_2},f_{2,z_2})&=(1,11),\\
r_3&=(s,t_3),&
z_3&=(s,u,v,w,x,t_3),&
(f_{3,r_3},f_{3,z_3})&=(3,7),\\
r_4&=(s,u,v,t_4),&
z_4&=(s,u,v,w,x,t_4),&
(f_{4,r_4},f_{4,z_4})&=(11,2).
\end{aligned}
\end{equation}
These are all source--terminal paths because every arc leaving the nonterminal chain enters a sink terminal. Positive unit costs were assigned only to the four shortcut arcs: $l_{(u,t_1)}=60$, $l_{(u,t_2)}=65$, $l_{(s,t_3)}=78$, and $l_{(v,t_4)}=60$. Every other unit arc cost was set to zero. These choices make the full-demand cost of each shortcut path equal to $780$.

The certificate is straightforward to verify. Let $\vy=(y_1,y_2,y_3,y_4)\in\{0,1\}^4$, where $y_i=1$ means that terminal $t_i$ uses $z_i$. The terminal-specific arcs always satisfy the additive load bound, and $(s,u)$ never binds. The remaining three shared arcs give
\begin{align}
&13y_1+12y_2+10y_3\leq34,\label{eq:minorfree-uv}\\
&13y_1+10y_3+13y_4\leq25,\label{eq:minorfree-vw}\\
&10y_3+13y_4\leq22.\label{eq:minorfree-wx}
\end{align}
Their minimal forbidden selections are $\{1,2,3\}$, $\{1,4\}$, and $\{3,4\}$. Enumerating the $16$ binary selections therefore gives
\begin{equation}\label{eq:minorfree-polytope}
\conv(\gY(\vf))
=
\left\{
\vy\in[0,1]^4:
y_1+y_4\leq1,\ 
y_3+y_4\leq1,\ 
y_1+y_2+y_3+y_4\leq2
\right\}.
\end{equation}
The fractional arborescence-path share vector is $\valpha=(3/13,11/12,7/10,2/13)$. It satisfies the two pairwise inequalities in Equation~\eqref{eq:minorfree-polytope}, whereas
\begin{equation}\label{eq:minorfree-facet-violation}
\sum_{i=1}^4\alpha_i = \frac{1561}{780} >2.
\end{equation}
The separating mechanism is therefore the higher-order cardinality facet $y_1+y_2+y_3+y_4\leq2$. The pairwise conflict graph contains only the edges $1$--$4$ and $3$--$4$ and is bipartite. The reference flow has cost $60\cdot10+65\cdot1+78\cdot3+60\cdot11=1559$. Every $1$-admissible selection uses at most two arborescence paths and hence at least two shortcut paths, each having full-demand cost $780$. The admissible selection $(1,1,0,0)$ attains this lower bound. Consequently,
\begin{equation}\label{eq:minorfree-cost-gap}
c_1^{\mathrm{u}}(\vf;\vl)
=
1560
>
1559
=
c(\vf;\vl).
\end{equation}

To further examine the reproducibility of the topology-transfer result and the contribution of the structural analysis, we conducted two additional searches in fresh conversations; Appendix~\ref{app:additional-ai-runs} reports their prompts and certificates. The first search repeated the transfer task with the full structural note, whereas the other one supplied only the basic problem setting and omitted the arborescence construction, shortcut reductions, and conflict-polytope framework developed in Sections~\ref{sec:preliminary}--\ref{sec:construction}. The outcomes differed substantially in their structural content. The full-note replication produced a distinct five-terminal counterexample satisfying the directed-topological-minor exclusion. By contrast, the basic-setting search returned a valid counterexample whose topology retained Rybin's structural core.

Overall, the results in this subsection show that the proposed pipeline is transferable beyond Rybin's seven-vertex topology. When supplied with the structural theory, the AI repeatedly constructed counterexamples on new topologies with higher-order conflict-polytope mechanisms, whereas the basic-setting searches tended to preserve Rybin's structural core.

\subsection{Testing the pipeline across varied initial arborescences}
\label{sec:systematic-tree-transfer}

The preceding experiments deliberately imposed topological novelty. We next removed that requirement and asked how broadly the same pipeline can succeed across distinct initial arborescences. In a single fresh conversation, ChatGPT was instructed to conduct a tree-first search. The core search instructions were as follows:
\begin{quote}
\emph{``Read the attached structural note carefully and use its forward construction pipeline to search systematically for counterexamples starting from many different arborescences. The objective is diversity of the initial arborescence, not novelty of the final augmented topology. The search must be tree-first. For each candidate, first generate and fix the initial arborescence; only then choose the shortcut arcs, demands, fractional path flows, and costs. Search among arborescences with three to five sink terminals and at most twelve vertices. Add exactly one terminal-entering ancestor shortcut for each terminal, so that every terminal has exactly two source--terminal paths. Require strictly positive fractional flow on both paths of every terminal and positive reference flow on every arc. Do not use inactive branches, redundant terminals, unused arcs, or inert subdivisions.
Before classifying a starting arborescence, suppress every nonterminal vertex with indegree one and outdegree one unless it is used as a shortcut tail. Two starting arborescences count as equivalent when their reduced forms are isomorphic under a mapping that preserves the source and terminal roles, permits terminal relabeling, and preserves the marked shortcut-tail vertices.
Find as many pairwise nonisomorphic starting-arborescence classes as possible."}
\end{quote}

\begin{figure}[ht]
\centering
\includegraphics[width=0.9\textwidth]{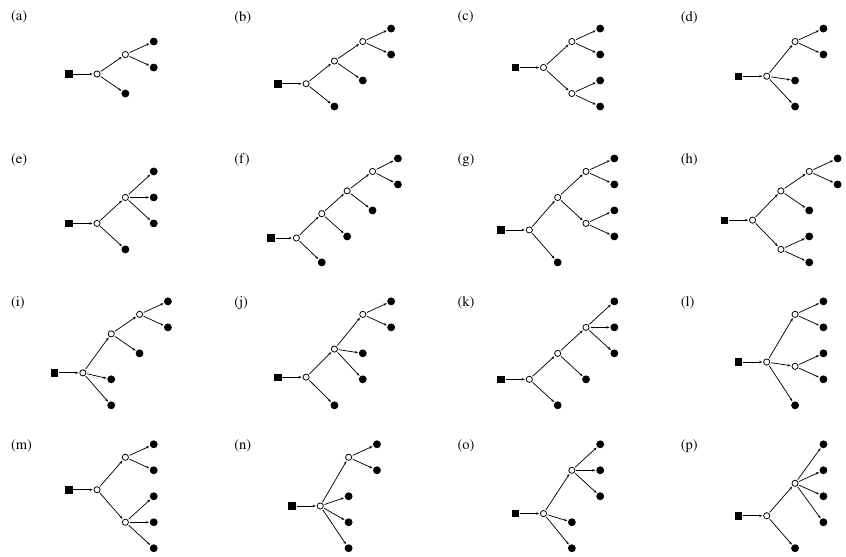}
\caption{The $16$ coarse branching skeletons represented among the successful starting arborescences. Each skeleton retains the source and terminals but suppresses every other vertex with indegree one and outdegree one, including shortcut tails, and forgets terminal labels. The filled square is the source, open circles are nonterminal branching vertices, filled circles are terminals, and arcs point away from the source.}
\label{fig:tree-skeletons}
\end{figure}

After approximately 62 minutes, ChatGPT produced $10{,}045$ nonisomorphic reduced marked starting classes. Grouping classes that induce the same family of choice-sensitive terminal sets yielded $572$ choice-set hypergraphs. A finite parameter search found exact counterexamples for $326$ of these hypergraphs, corresponding to $2{,}535$ starting classes. Independent exact verification checked $77{,}584$ unsplittable selections and $1{,}196{,}520$ arc-load inequalities without finding an error.  When shortcut-tail markings are forgotten, the successful classes span $94$ distinct rooted arborescence shapes: three with three terminals, $24$ with four terminals, and $67$ with five terminals. If every remaining nonterminal vertex with indegree one and outdegree one is also suppressed, including shortcut tails, these shapes reduce to the $16$ coarse branching skeletons in Figure~\ref{fig:tree-skeletons}: 1 with three terminals, 4 with four terminals, and 11 with five terminals. Appendix~\ref{app:systematic-tree-search} defines the different structural classes used in these counts and provides further details on the enumeration, parameter search, and exact verification.

This experiment demonstrates the breadth of the forward pipeline and ChatGPT's ability to operationalize the same construction principle across many different starting arborescences. The fact that the same procedure produced exactly verified counterexamples from 94 distinct rooted arborescence shapes shows that its success is not tied to the particular tree underlying Rybin's example. It also suggests a different view of the historical difficulty. Had this pipeline been formulated before the conjecture was resolved, we hypothesize that a human researcher could plausibly have applied it to small arborescences and found a counterexample within a relatively short time.

\section{Conclusion}
\label{sec:conclusion}

AI-generated and AI-assisted counterexamples are becoming a visible part of contemporary mathematical research, with recent examples spanning discrete geometry, probability, combinatorics, and operations research \citep{AlonEtAl2026UnitDistance,GuSellke2026,Liu2026ElekesRonyai,RamachandraNatarajan2026}. In this paper, we examine one such case in detail: Rybin's ChatGPT-generated seven-vertex counterexample to Goemans' unsplittable-flow conjecture.

By analyzing Rybin's counterexample, we developed a forward pipeline for constructing and testing further counterexamples. The pipeline begins with an arborescence, on which failure is impossible. It then introduces terminal-entering shortcuts incrementally, creating interacting route choices while keeping the complete set of graph-induced paths explicit. Finally, conflict-polytope analysis converts the resulting arc-load constraints into a finite geometric test: whether the fractional route-share vector lies outside the convex hull of the $1$-admissible unsplittable selections. This test then guides the construction of the demands, fractional path flows, and costs. The pipeline provides a forward constructive derivation of Tybin's counterexample. Further experiments demonstrate its breadth. First, when both the initial arborescence and the shortcut placement were allowed to change, ChatGPT constructed counterexamples on genuinely different topologies. Second, when the topology-novelty requirement was removed, a broad search produced exactly verified counterexamples from 94 distinct rooted arborescence shapes. These results show that the pipeline is not tied to the particular arborescence or conflict structure found in Rybin's example.

The same analysis explains why a counterexample to Goemans' conjecture remained difficult to discover. Verification begins only after the topology, complete path system, demands, fractional flow, and costs have all been fixed, whereas discovery must identify these components jointly. Even after the correct topology is supplied and the search is narrowed using substantial knowledge of the final answer, the counterexample-supporting region occupies only a minute fraction of the parameter space. The simplicity of the final counterexample therefore gives a misleading impression of how easily it could have been found.

For Rybin's interaction with GPT, our analysis places it between blind enumeration and the complete forward pipeline developed here. Based on the publicly available record, the model neither used nor established a forward procedure for identifying a topology capable of supporting a counterexample. The seven-vertex topology appeared only after broad computational exploration had eliminated many plausible constructions and yielded several partial structural restrictions. Once the topology appeared, ChatGPT imposed a symmetric parameterization and stated sufficient conditions encoding the three pair blockers and the triangle-facet violation. These conditions coincide with our Theorem~\ref{thm:complete-triangle} restricted to that symmetric subspace. They compressed the remaining search into a low-dimensional symbolic problem, which was crucial because our parameter-space analysis shows that brute-force search over the full space is almost impossible.

This case also points to a productive division of labor in mathematical research, organized as an iterative cycle between AI and human researchers. AI can move rapidly across cases, test variations, and adapt or extend existing theory, while human researchers can abstract from these trials, identify the important patterns, and turn them into general principles. With sustained interaction of this kind, mathematical theory may develop at a scale and speed that were previously difficult to imagine.

\bibliographystyle{apalike}
{\small
\bibliography{references}
}

\newpage
\appendix

\section{Proofs in Section \ref{sec:preliminary}}

\subsection{Proof of Proposition \ref{prop:two-shortcuts}}
\label{app:two-shortcuts}

If $\lvert\gA^{\mathrm{sc}}\rvert=0$, then $\gG=\gG^0$, and the result follows from Proposition~\ref{prop:unique-path}. Hence, suppose that $1\leq\lvert\gA^{\mathrm{sc}}\rvert\leq2$. For each terminal $t_i$ that receives a shortcut arc $(q_i,t_i)$, let $z_i$ denote its original $s$--$t_i$ path in $\gG^0$, and let $r_i$ denote the path formed by the unique $s$--$q_i$ path in $\gG^0$ followed by $(q_i,t_i)$. Define
\begin{equation}\label{eq:tree-route-share}
\alpha_i = \frac{\evf_{i,z_i}}{d_i}
\in[0,1].
\end{equation}
For each terminal $t_h$ that does not receive a shortcut arc, let $p_h$ denote its unique $s$--$t_h$ path in $\gG^0$. This is also the only $s$--$t_h$ path in $\gG$, so the reference flow necessarily satisfies $\evf_{h,p_h}=d_h$.

Suppose that $\lvert\gA^{\mathrm{sc}}\rvert=1$, and let $t_i$ be the terminal that receives the shortcut arc. Define two unsplittable flows $\vf^{z}$ and $\vf^{r}$ by $\evf^{z}_{i,z_i}=d_i$, $\evf^{r}_{i,r_i}=d_i$, and $\evf^{z}_{h,p_h} = \evf^{r}_{h,p_h} = d_h$ for every $h\neq i$; all remaining path-flow components of $\vf^{z}$ and $\vf^{r}$ are zero. Relative to the reference flow, either choice changes the load on any arc by at most $d_i\leq D$. Hence, both $\vf^z$ and $\vf^r$ are $1$-admissible. Moreover, as $\vf = \alpha_i \cdot \vf^z + (1-\alpha_i) \cdot \vf^r$, the linearity of the routing cost gives
\begin{equation}
c(\vf) = \alpha_i \cdot  c(\vf^z) + (1-\alpha_i) \cdot c(\vf^r),
\end{equation}
so at least one of $\vf^z$ and $\vf^r$ has cost no greater than $c(\vf)$.

Suppose now that $\lvert\gA^{\mathrm{sc}}\rvert=2$, and let $t_i$ and $t_j$ be the two terminals that receive shortcut arcs. Define four unsplittable flows $\vf^{zz}$, $\vf^{zr}$, $\vf^{rz}$, and $\vf^{rr}$ by
\begin{equation*}
\begin{aligned}
\evf^{zz}_{i,z_i}&=d_i,
&\evf^{zz}_{j,z_j}&=d_j,\\
\evf^{zr}_{i,z_i}&=d_i,
&\evf^{zr}_{j,r_j}&=d_j,\\
\evf^{rz}_{i,r_i}&=d_i,
&\evf^{rz}_{j,z_j}&=d_j,\\
\evf^{rr}_{i,r_i}&=d_i,
&\evf^{rr}_{j,r_j}&=d_j.
\end{aligned}
\end{equation*}
For every $h\notin\{i,j\}$, all four flows assign the entire demand $d_h$ to the unique path $p_h$, and all remaining path-flow components are zero. Thus, the first and second superscripts indicate the paths selected by terminals $t_i$ and $t_j$, respectively.

The flows $\vf^{rr}$, $\vf^{zr}$, and $\vf^{rz}$ are $1$-admissible. On a backbone arc, choosing a shortcut path cannot increase the reference load, while the only terminal that may choose its original path can increase the load by at most its demand, which does not exceed $D$. On an arc entering a terminal, the load increase is also at most the demand of that terminal. We then consider two cases:
\begin{itemize}
    \item Suppose first that $\vf^{zz}$ is $1$-admissible. The reference flow satisfies
\begin{equation}\label{eq:two-shortcut-square-decomposition}
\begin{aligned}
\vf=
\alpha_i\cdot \alpha_j\cdot\vf^{zz}
+\alpha_i\cdot (1-\alpha_j)\cdot \vf^{zr} +
(1-\alpha_i)\cdot \alpha_j\cdot \vf^{rz}
+(1-\alpha_i)\cdot (1-\alpha_j)\cdot \vf^{rr}.
\end{aligned}
\end{equation}
The coefficients are nonnegative and sum to one. By linearity of the routing cost, at least one of these four $1$-admissible unsplittable flows has cost no greater than $c(\vf)$.

\item Suppose instead that $\vf^{zz}$ is not $1$-admissible, and let $a$ be an arc on which it violates the load bound. Since $\vf^{zr}$ and $\vf^{rz}$ are $1$-admissible, the arc $a$ must be used by both original paths and bypassed by both shortcut paths; that is,
\begin{equation}
a \in
\left(
\gA_{i,z_i}\setminus\gA_{i,r_i}
\right)
\cap
\left(
\gA_{j,z_j}\setminus\gA_{j,r_j}
\right).
\end{equation}
On this arc, terminals $t_i$ and $t_j$ contribute $\alpha_i d_i+\alpha_jd_j$ in the reference flow and $d_i+d_j$ in $\vf^{zz}$, while all other contributions are unchanged. Hence,
\begin{equation}\label{eq:two-shortcut-blocker}
d_i+d_j
>
\alpha_i d_i+\alpha_jd_j+D.
\end{equation}
Without loss of generality, suppose that $d_i\geq d_j$. Since $D\geq d_i$, Equation \eqref{eq:two-shortcut-blocker} implies
\begin{equation}
(\alpha_i+\alpha_j)\cdot d_j
\leq
\alpha_i \cdot d_i+\alpha_j \cdot d_j < d_i+d_j-D \leq d_j.
\end{equation}
Therefore, $\alpha_i+\alpha_j<1$, and
\begin{equation}\label{eq:two-shortcut-simplex-decomposition}
\vf = \alpha_i\vf^{zr} + \alpha_j\vf^{rz} + (1-\alpha_i-\alpha_j)\cdot \vf^{rr}.
\end{equation}
The three coefficients are nonnegative and sum to one. Since $\vf^{zr}$, $\vf^{rz}$, and $\vf^{rr}$ are $1$-admissible, linearity of the routing cost implies that at least one of them has cost no greater than $c(\vf)$.
\end{itemize}

\subsection{Proof of Theorem~\ref{thm:shortcut-topology-uniqueness}}
\label{app:topology-uniqueness}

By Proposition~\ref{prop:two-shortcuts}, the assumed cost gap requires $\lvert\gA^{\mathrm{sc}}\rvert=3$. Since
Equation~\eqref{eq:shortcut-augmentation} allows at most one shortcut
arc entering each of the three terminals, exactly one arc is selected
from each of
$\gA_1^{\mathrm{sc}}$, $\gA_2^{\mathrm{sc}}$, and
$\gA_3^{\mathrm{sc}}$. It therefore remains to compare the
three-shortcut placements.

Label the three backbone arcs of $\gG^0$ by
$
b_1=(s,u)$, $b_2=(u,v)$, and $b_3=(v,w).
$
For each terminal $t_i$, let $(q_i,t_i)$ denote its shortcut arc, let $z_i$ denote its original $s$--$t_i$ path in $\gG^0$, and let $r_i$ denote the path formed by the unique $s$--$q_i$ path in $\gG^0$ followed by $(q_i,t_i)$. We record the backbone arcs that are used by $z_i$ and bypassed by $r_i$ through
\begin{equation}\label{eq:choice-interval}
I_i=\left\{j\in\{1,2,3\}:b_j\in\gA_{i,z_i}\setminus\gA_{i,r_i}\right\}.
\end{equation}
The allowable shortcut arcs give
\begin{equation}\label{eq:choice-interval-options}
I_1\in\{\{1,2\},\{2\}\}\quad\text{and}\quad I_i\in\{\{1,2,3\},\{2,3\},\{3\}\}\quad(i=2,3).
\end{equation}

Define
\begin{equation}\label{eq:appendix-tree-shares}
\alpha_i=\frac{\evf_{i,z_i}}{d_i}\in[0,1] \quad (i=1,2,3).
\end{equation}
For $y\in\{0,1\}^3$, let $\vf^{\mathrm{u}}(y)$ denote the unsplittable flow in which terminal $t_i$ uses $z_i$ when $y_i=1$ and $r_i$ when $y_i=0$. For each backbone arc $b_j$, define
\begin{equation}\label{eq:backbone-choice-set}
S_j=\{i\in\{1,2,3\}:j\in I_i\}.
\end{equation}
Thus, $S_j$ contains the terminals whose original paths use $b_j$ and whose shortcut paths bypass it.

\begin{lemma}[Admissibility on the backbone]\label{lem:backbone-admissibility}
The flow $\vf^{\mathrm{u}}(y)$ is $1$-admissible if and only if
\begin{equation}\label{eq:backbone-admissibility}
\sum_{i\in S_j}d_i\cdot y_i\leq\sum_{i\in S_j}\alpha_i\cdot d_i+D \quad (j=1,2,3).
\end{equation}
\end{lemma}

\begin{proof}
Fix $j\in\{1,2,3\}$. For each $i\in S_j$, terminal $t_i$ contributes $\alpha_i\cdot d_i$ to $b_j$ in the reference flow and $d_i\cdot y_i$ in $\vf^{\mathrm{u}}(y)$. For each $i\notin S_j$, its original and shortcut paths either both use $b_j$ or both avoid it, so its contribution is identical in the two flows. This gives \eqref{eq:backbone-admissibility}. Every arc outside the backbone enters one terminal and belongs to at most one of its two paths. Its load can therefore increase by at most that terminal's demand, which does not exceed $D$. Hence, the three inequalities in \eqref{eq:backbone-admissibility} completely characterize $1$-admissibility.
\end{proof}

The next lemma converts a weighted-average representation of the route shares into a cost guarantee.

\begin{lemma}\label{lem:convex-decomposition-certificate}

Let $\gY\subseteq\{0,1\}^3$ be the set of $1$-admissible selections for a fixed reference flow. Suppose that there exist $y^1,\ldots,y^N\in\gY$ and $\theta_1,\ldots,\theta_N\geq0$ satisfying
\begin{equation}\label{eq:alpha-convex-decomposition}
\sum_{l=1}^N\theta_l=1, \quad \alpha=\sum_{l=1}^N\theta_l\cdot y^l.
\end{equation}
Then, for every collection of nonnegative unit arc costs, at least one $1$-admissible unsplittable flow has cost no greater than the reference flow.
\end{lemma}

\begin{proof}
Let $\vf^{\mathrm{u},l}=\vf^{\mathrm{u}}(y^l)$. For each terminal $t_i$, the weighted average of these flows assigns
\begin{equation}
\sum_{l=1}^N\theta_l\cdot d_i\cdot y_i^l=\alpha_i\cdot d_i
\end{equation}
units to $z_i$ and the remaining $(1-\alpha_i)\cdot d_i$ units to $r_i$. Hence,
$
\vf=\sum_{l=1}^N\theta_l\cdot\vf^{\mathrm{u},l}.
$
Since the routing cost is linear in the path flows,
\begin{equation}
c(\vf)=\sum_{l=1}^N\theta_l\cdot c(\vf^{\mathrm{u},l}).
\end{equation}
At least one term in this weighted average is no greater than $c(\vf)$.
\end{proof}

We also use the following implication of an inadmissible pair selection.

\begin{lemma}\label{lem:blocked-pair}
Let $i,h,k$ be distinct indices in $\{1,2,3\}$. If the selection $y$ satisfying $y_i=y_h=1$ and $y_k=0$ is not $1$-admissible, then
\begin{equation}\label{eq:blocked-pair-share}
\alpha_i+\alpha_h<1.
\end{equation}
\end{lemma}

\begin{proof}
By Lemma~\ref{lem:backbone-admissibility}, the selection violates \eqref{eq:backbone-admissibility} on some backbone arc $b_j$. Its choice set $S_j$ must contain both $i$ and $h$; otherwise, the left-hand side of \eqref{eq:backbone-admissibility} is at most one terminal demand and hence at most $D$. Therefore,
\begin{equation}
d_i+d_h>\sum_{l\in S_j}\alpha_l\cdot d_l+D\geq\alpha_i\cdot d_i+\alpha_h\cdot d_h+D.
\end{equation}
Without loss of generality, suppose that $d_i\geq d_h$. Since $D\geq d_i$,
\begin{equation}
(\alpha_i+\alpha_h)\cdot d_h\leq\alpha_i\cdot d_i+\alpha_h\cdot d_h<d_i+d_h-D\leq d_h.
\end{equation}
Dividing by $d_h>0$ proves \eqref{eq:blocked-pair-share}.
\end{proof}

\subsubsection{Classification of the three-shortcut augmentations}

We abbreviate subsets of $\{1,2,3\}$ by concatenating their elements; for example, $12$ denotes $\{1,2\}$ and $123$ denotes $\{1,2,3\}$. Because $t_2$ and $t_3$ are symmetric in $\gG^0$, patterns that differ only by exchanging $I_2$ and $I_3$ belong to the same class. The interval options in \eqref{eq:choice-interval-options} produce the $12$ classes in Table~\ref{tab:three-shortcut-topologies}. The table also reports the choice sets $(S_1,S_2,S_3)$ from \eqref{eq:backbone-choice-set}.

\begin{table}[ht]
\centering
\small
\caption{Three-shortcut augmentations up to exchanging $t_2$ and $t_3$. The notation $12$ denotes $\{1,2\}$, and similarly for the other sets.}
\label{tab:three-shortcut-topologies}
\vspace{2pt}
\footnotesize
\begin{tabular}{@{}ccccc@{}}
\toprule
Pattern $(I_1,I_2,I_3)$ & $S_1$ & $S_2$ & $S_3$ & Classification \\
\midrule
$(12,123,123)$ & $123$ & $123$ & $23$ & Nested \\
$(12,123,23)$  & $12$  & $123$ & $23$ & Rybin \\
$(12,123,3)$   & $12$  & $12$  & $23$ & Pairwise \\
$(12,23,23)$   & $1$   & $123$ & $23$ & Nested \\
$(12,23,3)$    & $1$   & $12$  & $23$ & Pairwise \\
$(12,3,3)$     & $1$   & $1$   & $23$ & Pairwise \\
$(2,123,123)$  & $23$  & $123$ & $23$ & Nested \\
$(2,123,23)$   & $2$   & $123$ & $23$ & Nested \\
$(2,123,3)$    & $2$   & $12$  & $23$ & Pairwise \\
$(2,23,23)$    & $\varnothing$ & $123$ & $23$ & Nested \\
$(2,23,3)$     & $\varnothing$ & $12$  & $23$ & Pairwise \\
$(2,3,3)$      & $\varnothing$ & $1$   & $23$ & Pairwise \\
\bottomrule
\end{tabular}
\end{table}

We exclude the six pairwise classes and the five nested classes separately.

\subsubsection{Exclusion of the pairwise classes}

In a pairwise class, every choice set containing at least two terminals is either $\{1,2\}$ or $\{2,3\}$. Define a graph $H$ on $\{1,2,3\}$ by placing an edge $ih$ when the selection using $z_i$ and $z_h$ and the shortcut path of the remaining terminal is not $1$-admissible. The $1$-admissible selections are exactly the incidence vectors of the stable sets of $H$. Indeed, every selection containing an edge is inadmissible; the empty selection and all singleton selections are admissible; and a pair is admissible exactly when it is not an edge of $H$. If $H$ has no edges, the all-one selection also satisfies every backbone constraint because each nontrivial choice set contains only two terminals.

The graph $H$ is a subgraph of the path with edges $12$ and $23$, so it is bipartite. Lemma~\ref{lem:blocked-pair} gives $\alpha_i+\alpha_h<1$ for every edge $ih$ of $H$. The stable-set polytope of a bipartite graph is described by $0\leq y_i\leq1$ and $y_i+y_h\leq1$ for every edge $ih$ \citep{Chvatal1975,Schrijver2003}. Hence, $\alpha$ is a weighted average of the $1$-admissible selections, and Lemma~\ref{lem:convex-decomposition-certificate} excludes a counterexample in every pairwise class.

\subsubsection{Exclusion of the nested classes}

After repeated and singleton choice sets are removed, every nested class has at most the two nontrivial sets
\begin{equation}\label{eq:nested-choice-sets}
\{2,3\}\subseteq\{1,2,3\}.
\end{equation}
Let $H$ again denote the graph of blocked pair selections.

Suppose first that $H$ has one or two edges. The graph is bipartite, and its stable sets are exactly the $1$-admissible selections. Every selection containing an edge is inadmissible, while every stable set is the empty selection, a singleton, or a pair that is admissible by the definition of $H$. Lemma~\ref{lem:blocked-pair} gives every edge inequality satisfied by $\alpha$, so the bipartite stable-set description and Lemma~\ref{lem:convex-decomposition-certificate} exclude a counterexample.

Suppose next that $H$ has no edges. All three pair selections are then admissible. If $(1,1,1)$ is also admissible, every binary selection is admissible, and Lemma~\ref{lem:convex-decomposition-certificate} applies. If $(1,1,1)$ is inadmissible, it must be blocked by the constraint associated with $\{1,2,3\}$. Define $\xi_i=(1-\alpha_i)\cdot d_i$ for $i=1,2,3$. The inadmissibility of $(1,1,1)$ gives $\xi_1+\xi_2+\xi_3>D$, and hence
\begin{equation}
\sum_{i=1}^3(1-\alpha_i)=\sum_{i=1}^3\frac{\xi_i}{d_i}\geq\frac{\xi_1+\xi_2+\xi_3}{D}>1.
\end{equation}
Therefore, $\alpha_1+\alpha_2+\alpha_3<2$. The convex hull of the seven binary selections other than $(1,1,1)$ is $\{y\in[0,1]^3:y_1+y_2+y_3\leq2\}$, which contains $\alpha$. Lemma~\ref{lem:convex-decomposition-certificate} again excludes a counterexample.

Finally, suppose that $H$ has all three edges. Pairs $12$ and $13$ can be blocked only by the constraint associated with $\{1,2,3\}$. Define
$
A=\alpha_1\cdot d_1+\alpha_2\cdot d_2+\alpha_3\cdot d_3.
$
The two blocked pairs imply
\begin{equation}
A<d_1+d_2-D\leq\min\{d_1,d_2\}, \quad A<d_1+d_3-D\leq\min\{d_1,d_3\}.
\end{equation}
Thus, $A<d_{\min}$, where $d_{\min}=\min\{d_1,d_2,d_3\}$, and
\begin{equation}
\alpha_1+\alpha_2+\alpha_3=\sum_{i=1}^3\frac{\alpha_i\cdot d_i}{d_i}\leq\frac{A}{d_{\min}}<1.
\end{equation}
The admissible selections are then $(0,0,0)$ and the three singletons, whose convex hull is $\{y\in\mathbb{R}_+^3:y_1+y_2+y_3\leq1\}$. This set contains $\alpha$, and Lemma~\ref{lem:convex-decomposition-certificate} excludes a counterexample.

All $11$ non-Rybin classes in Table~\ref{tab:three-shortcut-topologies} have now been excluded. The remaining pattern is
$
(I_1,I_2,I_3)=(\{1,2\},\{1,2,3\},\{2,3\}),
$
with
$
(S_1,S_2,S_3)=(\{1,2\},\{1,2,3\},\{2,3\}).
$
By \eqref{eq:choice-interval-options}, this pattern is generated, up to exchanging $t_2$ and $t_3$, by
$
\gA^{\mathrm{sc}}=\{(s,t_1),(s,t_2),(u,t_3)\}.
$
This proves Theorem~\ref{thm:shortcut-topology-uniqueness}.

\section{Proofs in Section \ref{sec:construction}}\label{app:demand-proofs}

\subsection{Proof of Proposition~\ref{prop:canonical-cost}}\label{app:canonical-cost}

We use the path shares, selection vector, selected flow, and admissible selection set defined in Equations~\eqref{eq:general-reference-shares} and \eqref{eq:admissible-selection-set}. The set $\gY(\vf)$ is down-closed. To see this, take any $\vy\in\gY(\vf)$ and any coordinate $i$ with $y_i=1$. Changing $y_i$ from one to zero switches terminal $t_i$ from $z_i$ to $r_i$, weakly decreasing the load on every arc used by $z_i$ and bypassed by $r_i$. On the private shortcut arc of $r_i$, the unsplittable load becomes $d_i$, whereas the reference load is $(1-\alpha_i)d_i$. The increase is $\alpha_i d_i\leq d_i\leq D$. All other arc loads are unchanged or reduced. The resulting selection therefore remains in $\gY(\vf)$. Repeating this operation shows that if $\vy\in\gY(\vf)$ and $\bar\vy\leq\vy$ coordinatewise, then $\bar\vy\in\gY(\vf)$.

Suppose that $\vl\in\R_+^{|\gA|}$ satisfies $c_1^{\mathrm{u}}(\vf;\vl)>c(\vf;\vl)$. For each terminal $t_i$, define the costs of routing its full demand along $z_i$ and $r_i$ as
\begin{equation}
\kappa_i^z=d_i\cdot\sum_{a\in\gA_{i,z_i}}l_a
\quad\text{and}\quad
\kappa_i^r=d_i\cdot\sum_{a\in\gA_{i,r_i}}l_a,
\end{equation}
and let $\Delta_i=\kappa_i^r-\kappa_i^z$. For every $\vy\in\gY(\vf)$, the selected flow $\vf^{\mathrm{u}}(\vy)$ is $1$-admissible relative to $\vf$, so $c(\vf^{\mathrm{u}}(\vy);\vl)\geq c_1^{\mathrm{u}}(\vf;\vl)>c(\vf;\vl)$. Because the reference flow assigns $\alpha_i d_i$ units to $z_i$ and $(1-\alpha_i)d_i$ units to $r_i$, direct calculation gives
\begin{equation}\label{eq:canonical-proof-cost-gap}
c(\vf^{\mathrm{u}}(\vy);\vl)-c(\vf;\vl)
=
\sum_{i=1}^3\Delta_i\cdot(\alpha_i-y_i)
>0
\end{equation}
for every $\vy\in\gY(\vf)$.

Define $\Delta_i^+=\max\{\Delta_i,0\}$ and let $J_+=\{i:\Delta_i>0\}$. Fix any $\vy\in\gY(\vf)$, and define $\bar\vy\in\{0,1\}^3$ by $\bar y_i=y_i$ for $i\in J_+$ and $\bar y_i=0$ for $i\notin J_+$. Since $\bar\vy\leq\vy$, down-closedness gives $\bar\vy\in\gY(\vf)$. Applying Equation~\eqref{eq:canonical-proof-cost-gap} to $\bar\vy$ yields
\begin{equation}
0<\sum_{i=1}^3\Delta_i\cdot(\alpha_i-\bar y_i)=\sum_{i\in J_+}\Delta_i\cdot(\alpha_i-y_i)+\sum_{i\notin J_+}\Delta_i\cdot\alpha_i\leq\sum_{i=1}^3\Delta_i^+\cdot(\alpha_i-y_i).
\end{equation}
The last inequality uses $\Delta_i\leq 0$ and $\alpha_i\geq 0$ for every $i\notin J_+$. Since $\vy\in\gY(\vf)$ was arbitrary, $\sum_{i=1}^3\Delta_i^+\cdot(\alpha_i-y_i)>0$ for every $\vy\in\gY(\vf)$.

Set $K_i=\Delta_i^+$ for $i=1,2,3$, and define $\vl'=(l_a')_{a\in\gA}$ by
\begin{equation}
l_{(s,t_1)}'=\frac{K_1}{d_1},
\quad
l_{(s,t_2)}'=\frac{K_2}{d_2},
\quad
l_{(u,t_3)}'=\frac{K_3}{d_3},
\quad
l_a'=0
\quad
\text{for every }a\in\gA\setminus\gA^{\mathrm{p}}.
\end{equation}
Thus, $\vl'\in\gL'$. Under $\vl'$, every arborescence path $z_i$ has zero cost, while routing the full demand $d_i$ along shortcut path $r_i$ costs $K_i$. Consequently, for every $\vy\in\gY(\vf)$,
\begin{equation}
c(\vf^{\mathrm{u}}(\vy);\vl')-c(\vf;\vl')
=
\sum_{i=1}^3K_i\cdot(\alpha_i-y_i)
>0.
\end{equation}
The $1$-admissible unsplittable flows are precisely the flows $\vf^{\mathrm{u}}(\vy)$ generated by selections $\vy\in\gY(\vf)$. Moreover, $\gY(\vf)$ is finite and nonempty by Theorem~\ref{thm:dgg}. Taking the minimum over $\vy\in\gY(\vf)$ therefore gives
$
c_1^{\mathrm{u}}(\vf;\vl')>c(\vf;\vl'),
$
which proves the result.

\subsection{Proof of Theorem~\ref{thm:complete-triangle}}
\label{app:complete-triangle}

By Equations~\eqref{eq:general-reference-shares} and \eqref{eq:reference-path-flow}, $\valpha=(a/d_1,b/d_2,c/d_3)\in[0,1]^3$. For every selection $\vy\in\{0,1\}^3$, the loads of $\vf^{\mathrm{u}}(\vy)$ on the three shared backbone arcs $(s,u)$, $(u,v)$, and $(v,w)$ are $d_3+d_1\cdot y_1+d_2\cdot y_2$, $d_1\cdot y_1+d_2\cdot y_2+d_3\cdot y_3$, and $d_2\cdot y_2+d_3\cdot y_3$, respectively. The corresponding reference loads are $a+b+d_3$, $a+b+c$, and $b+c$. Every other arc is terminal-specific, so its unsplittable load can exceed its reference load by at most the corresponding terminal demand, which is no greater than $D$. Therefore, $\vy\in\gY(\vf)$ if and only if
\begin{align}
&d_1\cdot y_1+d_2\cdot y_2\leq a+b+D,\label{eq:selection-12}\\
&d_1\cdot y_1+d_2\cdot y_2+d_3\cdot y_3\leq a+b+c+D,\label{eq:selection-123}\\
&d_2\cdot y_2+d_3\cdot y_3\leq b+c+D.\label{eq:selection-23}
\end{align}
The left-hand sides are coordinatewise nondecreasing in $\vy$, so $\gY(\vf)$ is down-closed. The all-shortcut selection $(0,0,0)$ and the three selections using exactly one arborescence path, $(1,0,0)$, $(0,1,0)$, and $(0,0,1)$, always belong to $\gY(\vf)$ because the left-hand side of every applicable constraint is at most $D$.

Substituting $(1,1,0)$ into Equations~\eqref{eq:selection-12}--\eqref{eq:selection-23} shows that the pair $1$--$2$ is blocked exactly when $d_1+d_2>a+b+D$. Indeed, Equation~\eqref{eq:selection-12} is the strongest constraint for this selection: if it holds, then Equation~\eqref{eq:selection-123} holds because $c\geq0$, while Equation~\eqref{eq:selection-23} holds because $d_2\leq D$. Similarly, Equation~\eqref{eq:selection-123} is the only constraint that can block $(1,0,1)$, and Equation~\eqref{eq:selection-23} is the strongest constraint for $(0,1,1)$. Hence, the pairs $1$--$2$, $1$--$3$, and $2$--$3$ are blocked exactly when the first, second, and third inequalities in Equation~\eqref{eq:complete-system} hold, respectively.

\paragraph{Sufficiency.}
Suppose that the four inequalities in Equation~\eqref{eq:complete-system} hold. The exact blocker conditions established above imply that all three pairs are blocked. The four selections in Equation~\eqref{eq:triangle-selection-set} remain admissible. Every other selection uses at least two arborescence paths and is inadmissible, because otherwise down-closedness would make one of the blocked pair selections admissible. Hence, $\gY(\vf)=\gY_1$.

Choose any $K_0>0$ and define $\vl'$ by assigning unit costs $K_0/d_1$, $K_0/d_2$, and $K_0/d_3$ to $(s,t_1)$, $(s,t_2)$, and $(u,t_3)$, respectively, and zero unit cost to every other arc. Then $\vl'\in\gL'$, and each shortcut path has full-demand cost $K_0$. The fourth inequality gives $\alpha_1+\alpha_2+\alpha_3>1$, so Equation~\eqref{eq:equal-cost-gap} yields $c_1^{\mathrm{u}}(\vf;\vl')>c(\vf;\vl')$. Thus, a nonnegative unit arc-cost vector producing a counterexample exists.

\paragraph{Necessity.}
Suppose that $c_1^{\mathrm{u}}(\vf;\vl)>c(\vf;\vl)$ for some $\vl\in\R_+^{|\gA|}$. By Proposition~\ref{prop:canonical-cost}, there exists $\vl'\in\gL'$ such that $c_1^{\mathrm{u}}(\vf;\vl')>c(\vf;\vl')$. The convex-combination argument leading to Equation~\eqref{eq:outside-conflict-polytope} then gives $\valpha\notin\gC(\vf)$.
It remains to show that this condition requires all three pairs to be blocked.

Define a graph $H$ on $\{1,2,3\}$ by placing an edge between $i$ and $j$ whenever the pair $i$--$j$ is blocked. Every edge between $i$ and $j$ satisfies $\alpha_i+\alpha_j<1$. Indeed, the exact blocker conditions established above imply $d_i+d_j>\alpha_i d_i+\alpha_j d_j+D$; for the pair $1$--$3$, this follows after dropping the nonnegative term $b$ from the right-hand side. Without loss of generality, suppose that $d_i\geq d_j$. Since $D\geq d_i$,
\begin{equation}
(\alpha_i+\alpha_j)\cdot d_j\leq\alpha_i \cdot d_i+\alpha_j \cdot d_j<d_i+d_j-D\leq d_j,
\end{equation}
which proves $\alpha_i+\alpha_j<1$.

Suppose first that $H$ has one or two edges. The all-shortcut selection and the three singleton selections are admissible, and every unblocked pair selection is admissible by definition. The triple selection is inadmissible, because its admissibility and the down-closedness of $\gY(\vf)$ would imply that every pair selection is admissible. Hence, $\gY(\vf)$ consists exactly of the incidence vectors of the stable sets of $H$. Since $H$ is bipartite, its stable-set polytope is described by $0\leq y_i\leq1$ and $y_i+y_j\leq1$ for every edge between $i$ and $j$ \citep{Chvatal1975,Schrijver2003}. The vector $\valpha$ satisfies all these inequalities, so $\valpha\in\gC(\vf)$, a contradiction.

Suppose next that $H$ has no edges. All three pair selections are admissible. If $(1,1,1)$ is also admissible, then $\gY(\vf)=\{0,1\}^3$, so $\gC(\vf)=[0,1]^3$ contains $\valpha$. If $(1,1,1)$ is inadmissible, it can violate only Equation~\eqref{eq:selection-123}. Indeed, Equations~\eqref{eq:selection-12} and \eqref{eq:selection-23} have the same left-hand sides under $(1,1,1)$ as under the admissible pair selections $(1,1,0)$ and $(0,1,1)$, respectively. The inadmissibility of $(1,1,1)$ therefore gives $(d_1-a)+(d_2-b)+(d_3-c)>D$. Since $d_i\leq D$,
\begin{equation}
\sum_{i=1}^3(1-\alpha_i)=\frac{d_1-a}{d_1}+\frac{d_2-b}{d_2}+\frac{d_3-c}{d_3}\geq\frac{(d_1-a)+(d_2-b)+(d_3-c)}{D}>1.
\end{equation}
Thus, $\alpha_1+\alpha_2+\alpha_3<2$. In this case, $\gY(\vf)$ contains every binary selection except $(1,1,1)$, and
\begin{equation}
\gC(\vf)=[0,1]^3\cap\{\vy\in\R^3:y_1+y_2+y_3\leq2\}.
\end{equation}
This set contains $\valpha$, again a contradiction.

Therefore, $H$ contains all three edges. By the exact blocker conditions established above, the first three inequalities in Equation~\eqref{eq:complete-system} are necessary. Moreover, all three blocked pairs give $\gY(\vf)=\gY_1$, and hence
\begin{equation}
\gC(\vf)=\{\vy\in\R_+^3:y_1+y_2+y_3\leq1\}.
\end{equation}
Because $\valpha\notin\gC(\vf)$ and $\valpha\in[0,1]^3$, we must have $\alpha_1+\alpha_2+\alpha_3>1$. Since $\alpha_1=a/d_1$, $\alpha_2=b/d_2$, and $\alpha_3=c/d_3$, this is the fourth inequality in Equation~\eqref{eq:complete-system}. This completes the proof.

\subsection{Proof of Theorem~\ref{thm:demand-region}}
\label{app:demand-region-proof}

Normalize all demands and the arborescence-path flows $a,b,c$ by $D$, and write $A=a/D$, $B=b/D$, and $C=c/D$. The normalized demand values are $1\geq\sigma\geq\rho>0$. By Theorem~\ref{thm:complete-triangle}, a counterexample exists exactly when the three blocker inequalities admit a fractional flow whose total arborescence-path share exceeds one. We first replace the three strict blocker inequalities by their non-strict versions and maximize this total share over the resulting closed region.

We begin by determining which demand can be assigned to the central terminal $t_2$. Suppose first that $d_2=M$. By endpoint symmetry, it suffices to consider the normalized assignment
\begin{equation}
\left(\frac{d_1}{D},\frac{d_2}{D},\frac{d_3}{D}\right)=(1,\sigma,\rho).
\end{equation}
The non-strict blocker for the pair $1$--$3$ gives $A+B+C\leq\rho$. Since $1/\rho$ is the largest reciprocal demand,
\begin{equation}
A+\frac{B}{\sigma}+\frac{C}{\rho}\leq\frac{A+B+C}{\rho}\leq1.
\end{equation}
Thus, the total arborescence-path share cannot exceed one.

Suppose next that $d_2=D$. By endpoint symmetry, consider
\begin{equation}
\left(\frac{d_1}{D},\frac{d_2}{D},\frac{d_3}{D}\right)=(\sigma,1,\rho).
\end{equation}
The non-strict blockers for the pairs $1$--$3$ and $2$--$3$ imply $A+B+C\leq\sigma+\rho-1$ and $B+C\leq\rho$.
Multiplying the first inequality by $1/\sigma$ and the second by $1/\rho-1/\sigma\geq0$ gives an inequality whose left-hand side dominates the total share. Hence,
\begin{equation}
\frac{A}{\sigma}+B+\frac{C}{\rho}\leq\frac{\sigma+\rho-1}{\sigma}+\rho\cdot\left(\frac{1}{\rho}-\frac{1}{\sigma}\right)=2-\frac{1}{\sigma}\leq1.
\end{equation}
Therefore, a counterexample requires $d_2=S$. By endpoint symmetry, it suffices to analyze the normalized assignment
\begin{equation}
\left(\frac{d_1}{D},\frac{d_2}{D},\frac{d_3}{D}\right)=(1,\rho,\sigma).
\end{equation}

Under this assignment, the closed blocker region is
\begin{align}
&A+B\leq\rho,\label{eq:LP1}\\
&A+B+C\leq\sigma,\label{eq:LP2}\\
&B+C\leq\rho+\sigma-1,\label{eq:LP3}\\
&A,B,C\geq0.\label{eq:LP4}
\end{align}
The total arborescence-path share is
\begin{equation}\label{eq:Phi-objective}
\Phi(A,B,C)=A+\frac{B}{\rho}+\frac{C}{\sigma}.
\end{equation}
If $\rho+\sigma<1$, the region is empty by Equation~\eqref{eq:LP3}. Hence, assume $\rho+\sigma\geq1$.

Suppose first that $\rho\leq1/2$, and define
$
\theta=\max\left\{1 / \sigma, 1 / \rho-1\right\}.
$
Adding Equation~\eqref{eq:LP1} to $\theta$ times Equation~\eqref{eq:LP3} gives an inequality whose coefficients dominate those in Equation~\eqref{eq:Phi-objective}. Therefore,
\begin{equation}
\Phi(A,B,C)\leq\rho+\theta\cdot(\rho+\sigma-1)\leq1.
\end{equation}
For $\theta=1/\sigma$, the last inequality follows from $\rho\cdot\sigma\leq\rho\leq1-\rho$; for $\theta=1/\rho-1$, its right-hand side equals $1+(1-\rho)\cdot(\sigma-1)/\rho\leq1$. Thus, no counterexample exists when $\rho\leq1/2$.

Suppose now that $\rho>1/2$. The point
\begin{equation}\label{eq:ABC-star}
A^*=1-\rho,
\quad
B^*=2\rho-1,
\quad
C^*=\sigma-\rho
\end{equation}
satisfies Equations~\eqref{eq:LP1}--\eqref{eq:LP3} at equality. Consider the multipliers
\begin{equation}\label{eq:dual-multipliers}
\lambda_1=\frac{1}{\rho}-\frac{1}{\sigma},
\quad
\lambda_2=1+\frac{1}{\sigma}-\frac{1}{\rho},
\quad
\lambda_3=\frac{1}{\rho}-1.
\end{equation}
They are nonnegative because $\rho\leq\sigma\leq1$ and $\rho>1/2$. Multiplying Equations~\eqref{eq:LP1}, \eqref{eq:LP2}, and \eqref{eq:LP3} by $\lambda_1$, $\lambda_2$, and $\lambda_3$, respectively, reproduces the coefficients in Equation~\eqref{eq:Phi-objective}. The resulting upper bound is attained at Equation~\eqref{eq:ABC-star}. Hence,
\begin{equation}
\Phi^*(\rho,\sigma):=\max\Phi
=
4-\rho-\frac{1}{\rho}-\frac{\rho}{\sigma}.
\end{equation}
A total share greater than one is therefore possible exactly when $\rho>1/2$ and $\Phi^*(\rho,\sigma)>1$. Since $3\rho-\rho^2-1>0$ for $\rho>1/2$, rearranging the latter inequality gives
$\sigma>\rho^2 / (3\rho-\rho^2-1).$
Because $\sigma\leq1$, this strict inequality also requires $\rho<1$. This proves the necessity of Equation~\eqref{eq:demand-region}.

Conversely, suppose that Equation~\eqref{eq:demand-region} holds. Then $\Phi^*(\rho,\sigma)>1$, and
\begin{equation}
\sigma>\frac{\rho^2}{3\rho-\rho^2-1}>\rho.
\end{equation}
Thus, all three coordinates in Equation~\eqref{eq:ABC-star} are positive and smaller than the corresponding normalized demands. Choose any $\eta$ satisfying $1/\Phi^*(\rho,\sigma)<\eta<1$, and set
\begin{equation}
A=\eta\cdot(1-\rho),
\quad
B=\eta\cdot(2\rho-1),
\quad
C=\eta\cdot(\sigma-\rho).
\end{equation}
Because Equations~\eqref{eq:LP1}--\eqref{eq:LP3} are tight at $(A^*,B^*,C^*)$, multiplying this point by $\eta<1$ makes all three blocker inequalities strict. Moreover,
$\Phi(A,B,C)=\eta \cdot \Phi^*(\rho,\sigma)>1.$
Rescaling by $D$ gives
\begin{equation}
a=\eta\cdot(D-S),
\quad
b=\eta\cdot(2S-D),
\quad
c=\eta\cdot(M-S).
\end{equation}
These values satisfy all four inequalities in Equation~\eqref{eq:complete-system}. Theorem~\ref{thm:complete-triangle} therefore gives a fractional flow and a nonnegative unit arc-cost vector producing a counterexample for $(d_1,d_2,d_3)=(D,S,M)$. Exchanging the endpoint terminals gives the same conclusion for $(d_1,d_2,d_3)=(M,S,D)$. This completes the proof.

\section{Additional AI-search runs}
\label{app:additional-ai-runs}

We supplemented the experiments in Section~\ref{sec:ai-search} with two fresh-conversation runs. The first repeated the stricter topology-transfer experiment with the full structural note. The second supplied only the basic-setting note, where the results in Sections \ref{sec:preliminary}-- \ref{sec:construction} are not included.

\subsection{A five-terminal replication with the structural note}
\label{app:five-terminal-replication}

\begin{figure}[ht]
\centering
\includegraphics[width=0.6\textwidth]{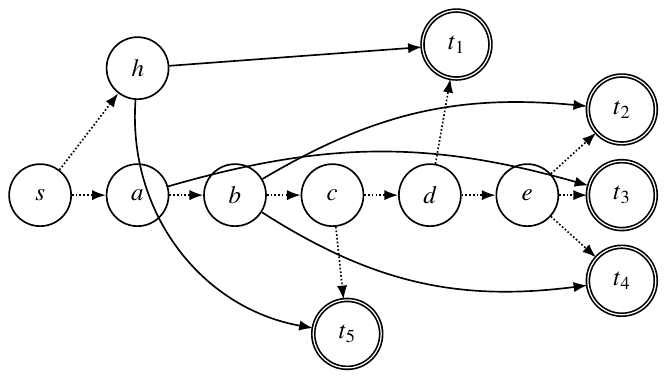}
\caption{Five-terminal topology from an additional theory-guided run. Dotted arcs are zero-cost; solid arcs are positive-cost shortcuts.}
\label{fig:five-terminal-replication}
\end{figure}

Repeating the stricter prompt from Section~\ref{sec:transfer-pipeline} with the full structural note produced the five-terminal topology in Figure~\ref{fig:five-terminal-replication}. Its vertex set is
$\{s,h,a,b,c,d,e,t_1,t_2,t_3,t_4,t_5\}$. The zero-cost arborescence arcs are
\begin{equation}
\begin{aligned}
\gA^{0}_{\mathrm{rep}}=\{(s,a),(a,b),(b,c),(c,d),(d,e),(s,h),(d,t_1),(e,t_2),(e,t_3),(e,t_4),(c,t_5)\},
\end{aligned}
\end{equation}
and the five positive-cost terminal-entering shortcuts are
\begin{equation}
\gA^{\mathrm{p}}_{\mathrm{rep}}
=
\{(h,t_1),(b,t_2),(a,t_3),(b,t_4),(h,t_5)\}.
\end{equation}

Each terminal has exactly two source--terminal paths:
\begin{equation*}
\begin{aligned}
r_1&=(s,h,t_1),&
 z_1&=(s,a,b,c,d,t_1),\quad
&(f_{1,r_1},f_{1,z_1})=(19,1),\\
r_2&=(s,a,b,t_2),&
 z_2&=(s,a,b,c,d,e,t_2),
&(f_{2,r_2},f_{2,z_2})=(1,19),\\
r_3&=(s,a,t_3),&
 z_3&=(s,a,b,c,d,e,t_3),
&(f_{3,r_3},f_{3,z_3})=(10,10),\\
r_4&=(s,a,b,t_4),&
 z_4&=(s,a,b,c,d,e,t_4),
&(f_{4,r_4},f_{4,z_4})=(19,5),\\
r_5&=(s,h,t_5),&
 z_5&=(s,a,b,c,t_5),
&(f_{5,r_5},f_{5,z_5})=(16,8).
\end{aligned}
\end{equation*}
The demand vector is
$
(d_1,d_2,d_3,d_4,d_5)=(20,20,20,24,24),
$
so $D=24$. These are all source--terminal paths because the nonterminal subgraph is an arborescence and every arc leaving it enters a sink terminal.

Positive unit costs are assigned only to the shortcut arcs:
\begin{equation}
 l_{(h,t_1)}=l_{(b,t_2)}=l_{(a,t_3)}=\frac{1}{20},
 \quad
 l_{(b,t_4)}=l_{(h,t_5)}=\frac{1}{24}.
\end{equation}
Every full-demand shortcut path therefore costs one, while every arborescence path has zero cost. Let $\vy\in\{0,1\}^5$, where $y_i=1$ means that terminal $t_i$ uses $z_i$. The five backbone arcs give the exact choice-sensitive inequalities
\begin{align*}
&20y_1+24y_5\leq33,\\
&20y_1+20y_3+24y_5\leq43,\\
&20y_1+20y_2+20y_3+24y_4+24y_5\leq67,\\
&20y_1+20y_2+20y_3+24y_4\leq59,\\
&20y_2+20y_3+24y_4\leq58.
\end{align*}
The remaining arcs automatically satisfy the additive load bound. Enumerating the $32$ binary selections gives
\begin{equation}\label{eq:replication-selection-set}
\begin{aligned}
\gY(\vf)=\{0,e_1,e_2,e_3,e_4,e_5,
 e_1+e_2,e_1+e_3,e_1+e_4, e_2+e_3,e_2+e_4,e_2+e_5,e_3+e_4,e_4+e_5\},
\end{aligned}
\end{equation}
where $e_i$ denotes the $i$th unit vector. In particular, every admissible selection satisfies
\begin{equation}\label{eq:replication-cardinality-facet}
\sum_{i=1}^5y_i\leq2.
\end{equation}
This inequality is a facet of $\conv(\gY(\vf))$. The pair-conflict graph has only the edges $1$--$5$ and $3$--$5$, so it contains no odd cycle. The fractional arborescence-path share vector is
\begin{equation}
\valpha=\left(\frac1{20},\frac{19}{20},\frac12,\frac5{24},\frac13\right),
\quad
\sum_{i=1}^5\alpha_i=\frac{49}{24}>2,
\end{equation}
while $\alpha_1+\alpha_5=23/60<1$ and $\alpha_3+\alpha_5=5/6<1$. Separation therefore comes from the five-coordinate cardinality facet rather than from the pair-conflict graph.

The reference cost is
\begin{equation}
 c(\vf;\vl)
 =\frac{19}{20}+\frac{1}{20}+\frac{10}{20}+\frac{19}{24}+\frac{16}{24}
 =\frac{71}{24}.
\end{equation}
Every $1$-admissible unsplittable flow uses at least three shortcut paths, and an admissible pair selection attains this bound. Hence,
\begin{equation}\label{eq:replication-cost-gap}
 c_1^{\mathrm{u}}(\vf;\vl)
 =3
 >
 \frac{71}{24}
 =c(\vf;\vl).
\end{equation}
The source has the two outgoing arcs $(s,a)$ and $(s,h)$. The allowed deletion and suppression operations cannot increase source outdegree, whereas Rybin's source has outdegree three. The topology therefore satisfies the source-preserving directed-topological-minor exclusion used in the prompt. This independent run thus produced a different topology and a five-dimensional realization of the same higher-order rank-two mechanism observed in Section~\ref{sec:transfer-pipeline}.

\subsection{A basic-setting control}
\label{app:basic-setting-controls}

To understand the contribution of the structural analysis, we conducted a control in a separate fresh conversation. The conversation received only the basic-setting note (Section \ref{sec:model}), which states the problem and records Rybin's counterexample but omits the arborescence construction, shortcut reductions, and conflict-polytope framework developed in Sections~\ref{sec:preliminary}--\ref{sec:construction}. It then used the following prompt:
\vspace{5pt}
\begin{quote}
\itshape
``Read the attached material carefully and search for a complete counterexample on a topology that does not contain Rybin's seven-vertex network as a directed topological minor. Concretely, for no choice of three terminals may Rybin's topology be obtained, up to relabeling that preserves the source and terminal roles, by deleting vertices or arcs and then suppressing nonterminal vertices with indegree one and outdegree one in the remaining graph. Do not return an active extension of Rybin's network.
The counterexample may use any number of terminals. Give the graph, terminal demands, fractional path flows, and nonnegative arc costs. Enumerate all source--terminal paths, verify exactly that every $1$-admissible unsplittable flow has strictly greater cost than the fractional reference flow, and provide a graph-theoretic proof that the topology satisfies the directed-topological-minor exclusion above. Think for no more than 75 minutes. Use only the attached material and the information provided in this conversation."
\end{quote}
\vspace{5pt}

\begin{figure}[ht]
\centering
\includegraphics[width=0.5\textwidth]{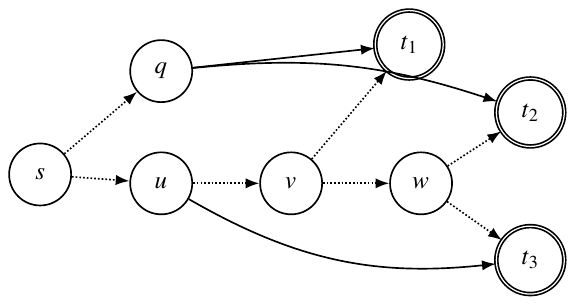}
\caption{Topology returned by the first basic-setting control. Dotted arcs have zero unit cost and solid arcs have positive unit cost. Contracting $(s,q)$ recovers Rybin's topology in Figure~\ref{fig:rybin-certificate} exactly.}
\label{fig:basic-control-source-split}
\end{figure}

The response returned the eight-vertex graph in Figure~\ref{fig:basic-control-source-split}. Relative to Rybin's topology, it replaced the two arcs $(s,t_1)$ and $(s,t_2)$ by the three-arc branch
\begin{equation}\label{eq:basic-control-source-split}
(s,t_1),(s,t_2)
\quad\longrightarrow\quad
(s,q),(q,t_1),(q,t_2),
\end{equation}
and left all other arcs unchanged.

The response also reused Rybin's demands, fractional path flows, and costs on the corresponding arcs; every other arc had zero unit cost. Consequently, the admissible-selection set and cost gap were identical to Rybin's:
\begin{equation}\label{eq:basic-control-certificate}
\gY(\vf)
=
\{(0,0,0),(1,0,0),(0,1,0),(0,0,1)\},
\quad
c_1^{\mathrm{u}}(\vf;\vl)=60>58=c(\vf;\vl).
\end{equation}
The returned graph passed the literal source-preserving directed-topological-minor test because its designated source has outdegree two and $q$ is not suppressible under the indegree-one/outdegree-one rule. However, it did not produce a substantively new topology. This control suggests that, without an explicit framework for organizing the search, ChatGPT may find it difficult to move beyond the structural template of the known counterexample.

\section{Details of the systematic arborescence search}
\label{app:systematic-tree-search}

Section~\ref{sec:systematic-tree-transfer} reports the principal findings of the search. This appendix defines the structural classes used in those counts, explains their reduction to choice-set hypergraphs, and summarizes the parameter search and exact verification.

The structural search began with undirected tree shapes having at most twelve vertices. Each tree was rooted at every possible source, with all arcs directed away from the root. Retained arborescences had three to five sink terminals, each at depth at least two. For every terminal $t_i$, exactly one strict ancestor of its parent was selected as the tail of a shortcut entering $t_i$. A nonterminal vertex with indegree one and outdegree one was suppressed unless it served as a shortcut tail. Two resulting objects were identified when a rooted-tree isomorphism preserved the source, mapped terminals to terminals, permitted terminal relabeling, and preserved the corresponding shortcut-tail locations. An equivalence class under these rules is called a \emph{reduced marked starting-arborescence class}. The marking is essential: a single unmarked rooted tree can generate several classes when its shortcuts depart from inequivalent vertices. Within the finite scope specified in the prompt, the structural enumeration produced $10{,}045$ such classes.

For each tree arc $e$, define its choice-sensitive terminal set by
\[
S_e
=
\left\{
i:
e\in A_{i,z_i}
\ \text{and}\
e\notin A_{i,r_i}
\right\},
\]
where $z_i$ is terminal $t_i$'s arborescence path and $r_i$ is its shortcut path. Continue to write
$
a_i=f_{i,z_i}
$
and
$
\alpha_i=a_i/d_i.
$
If $y_i=1$ means that terminal $t_i$ uses $z_i$, then, after cancelling route-independent loads, the $1$-admissibility condition on $e$ is
\begin{equation}\label{eq:systematic-choice-set-bound}
\sum_{i\in S_e} d_i y_i
\leq
\sum_{i\in S_e} a_i+D.
\end{equation}
Empty, singleton, and duplicate sets can be omitted because they add no nontrivial restriction. The remaining family of sets, considered up to terminal relabeling, is the \emph{choice-set hypergraph} of the starting class.

Starting classes with the same choice-set hypergraph induce the same binary admissibility system in Equation~\eqref{eq:systematic-choice-set-bound}. A demand-and-flow construction found for one hypergraph can therefore be transferred to every marked arborescence realizing that hypergraph, after matching the terminal labels. This reduction compresses the $10{,}045$ starting classes into $572$ distinct hypergraph types.

For each hypergraph, the parameter search generated positive demands and positive path flows and enumerated all $2^k$ binary selections exactly. Every retained counterexample could be certified by a uniform rank inequality
\begin{equation}\label{eq:systematic-uniform-rank-facet}
\sum_{i=1}^k y_i\leq r,
\qquad
(k,r)\in
\{(3,1),(4,1),(4,2),(5,1),(5,2),(5,3)\},
\end{equation}
that was violated by the fractional share vector:
$
\sum_i\alpha_i>r.
$
Assigning unit cost $1/d_i$ to terminal $t_i$'s private shortcut arc and zero cost to every tree arc makes the full-demand cost of each shortcut path equal to one. The reference cost is then
$$
c(\vf;\vl) = k-\sum_{i=1}^k\alpha_i,
$$
whereas every $1$-admissible unsplittable flow has cost at least $k-r$.
Thus, the violation of Equation~\eqref{eq:systematic-uniform-rank-facet} yields the strict cost gap directly.

The search found exact counterexamples for $326$ choice-set hypergraphs, corresponding to $2{,}535$ reduced marked starting classes. The structural distribution is summarized in Table~\ref{tab:systematic-search-outcomes}.

\begin{table}[ht]
\centering
\caption{Structural classes represented among the verified counterexamples.}
\label{tab:systematic-search-outcomes}
\scriptsize
\setlength{\tabcolsep}{4pt}
\begin{tabular}{@{}c
r
r
r
r
r
r@{}}
\toprule
\shortstack{Number of\\terminals}
&
\shortstack{Catalogued\\marked classes}
&
\shortstack{Catalogued\\hypergraphs}
&
\shortstack{Successful\\hypergraphs}
&
\shortstack{Successful\\marked classes}
&
\shortstack{Rooted tree\\shapes}
&
\shortstack{Coarse\\skeletons}
\\
\midrule
$3$ & $44$      & $5$   & $1$   & $6$       & $3$  & $1$  \\
$4$ & $782$     & $39$  & $19$  & $212$     & $24$ & $4$  \\
$5$ & $9{,}219$ & $528$ & $306$ & $2{,}317$ & $67$ & $11$ \\
\midrule
Total & $10{,}045$ & $572$ & $326$ & $2{,}535$ & $94$ & $16$ \\
\bottomrule
\end{tabular}
\end{table}

These counts describe three levels of structural resolution. The $2{,}535$ marked classes distinguish both the rooted tree and the shortcut-tail locations. Forgetting the shortcut-tail markings leaves $94$ rooted arborescence shapes. Suppressing all remaining nonterminal vertices with indegree one and outdegree one, including shortcut tails, gives the $16$ coarse branching skeletons displayed in Figure~\ref{fig:tree-skeletons}. The $2{,}535$ final augmented networks are also pairwise nonisomorphic under directed isomorphisms preserving the source and terminal roles; exactly one is isomorphic to Rybin's topology.
Every retained construction satisfies the activity conditions imposed in the prompt. Each terminal has exactly two source--terminal paths and sends positive fractional flow on both; every tree arc and shortcut carries positive reference flow; and no retained degree-two nonterminal is an inert subdivision. The constructions also passed a terminal-nonredundancy test, so the four- and five-terminal examples are not obtained by adding inactive terminals to smaller counterexamples.
A separate exact verifier reconstructed every retained graph and checked every binary selection arc by arc. Across all $2{,}535$ constructions, it examined $77{,}584$ unsplittable selections and $1{,}196{,}520$ arc-load inequalities, and found no error. Reference costs, minimum admissible unsplittable costs, and strict gaps were recomputed using exact rational arithmetic.

The scope of the conclusions requires one qualification. The enumeration of the $10{,}045$ reduced marked starting classes is exhaustive under the stated limits on vertices, terminals, shortcut number, and reducedness. The parameter search is a large but finite search over a continuous space and is not exhaustive. Hence, the $2{,}535$ successful marked classes, $94$ rooted tree shapes, and $16$ coarse skeletons are certified lower bounds on the breadth of the pipeline. The six rank-facet types in Equation~\eqref{eq:systematic-uniform-rank-facet} are likewise the mechanisms found by the search, rather than an exhaustive classification of all facets that these arborescences may support.

\end{document}